\documentclass[runningheads]{llncs}

\usepackage{eccv}

\usepackage{eccvabbrv}

\usepackage{graphicx}
\usepackage{booktabs}
\usepackage{epsfig}
\usepackage{multicol}
\usepackage{multirow}
\usepackage{wrapfig}
\usepackage{marvosym}
\usepackage{array}
\usepackage{pifont}
\newcommand{\tcr}{\textcolor{red}}
\usepackage[accsupp]{axessibility}  

\usepackage{hyperref}

\usepackage{orcidlink}

\begin{document}


\title{UCIP: A Universal Framework for Compressed Image Super-Resolution using Dynamic Prompt}

\titlerunning{UCIP}

\author{Xin Li\inst{1}\orcidlink{0000-0002-6352-6523}\textsuperscript{$\dagger$} \and
Bingchen Li\inst{1}\orcidlink{0009-0001-9990-7790}\textsuperscript{$\dagger$} \and
Yeying Jin\inst{2}\orcidlink{0000-0001-7818-9534} \and
Cuiling	Lan\inst{3}\orcidlink{0000-0001-9145-9957} \and
Hanxin Zhu\inst{1}\orcidlink{0009-0006-3524-0364}  \\ 
Yulin Ren\inst{1}\orcidlink{0009-0006-4815-7973} \and
Zhibo Chen\inst{1}\orcidlink{0000-0002-8525-5066}}

\authorrunning{X.~Li et al.}

\institute{University of Science and Technology of China \and
National University of Singapore \and
Microsoft Research Asia\\
\small \email{\{xin.li, chenzhibo\}@ustc.edu.cn, 
\{lbc31415926, hanxinzhu, renyulin\}@mail.ustc.edu.cn, jinyeying@u.nus.edu, culan@microsoft.com}}

\maketitle
\renewcommand{\thefootnote}{}
\footnotetext{\textsuperscript{$\dagger$}~Equal Contribution.}

\begin{abstract}
Compressed Image Super-resolution (CSR) aims to simultaneously super-resolve the compressed images and tackle the challenging hybrid distortions caused by compression. However, existing works on CSR usually focus on single compression codec, \ie, JPEG, ignoring the diverse traditional or learning-based codecs in the practical application, \eg, HEVC, VVC, HIFIC, \emph{etc}. In this work, we propose the first universal CSR framework, dubbed UCIP, with dynamic prompt learning, intending to jointly support the CSR distortions of any compression codecs/modes. Particularly, an efficient dynamic prompt strategy is proposed to mine the content/spatial-aware task-adaptive contextual information for the universal CSR task, using only a small amount of prompts with spatial size $1\times1$. To simplify contextual information mining, we introduce the novel MLP-like framework backbone for our UCIP by adapting the Active Token Mixer (ATM) to CSR tasks for the first time, where the global information modeling is only taken in horizontal and vertical directions with offset prediction. We also build an all-in-one benchmark dataset for the CSR task by collecting the datasets with the popular 6 diverse traditional and learning-based codecs, including JPEG, HEVC, VVC, HIFIC, etc., resulting in 23 common degradations. Extensive experiments have shown the consistent and excellent performance of our UCIP on universal CSR tasks. The project can be found in~\textcolor{magenta}{\url{https://lixinustc.github.io/UCIP.github.io}}


\end{abstract}

\section{Introduction}
\label{sec:intro}
In recent years, we have witnessed the significant development of Deep Neural Networks (DNNs) in image super-resolution (SR)~\cite{li2023cswin2sr,wang2018esrganRRDB,fritsche2019freqRealSR,li2022hst,realesrgan,RCAN,stablesr,msgdn,wu2023seesr,sun2023coser}, where the image is degraded with low-resolution artifacts. However, in the practical scenario, due to the limitation of storage and bandwidth, collected images are also inevitably compressed with traditional image codecs, such as JPEG~\cite{JPEG}, and BPG~\cite{HM}. Accordingly, compressed image super-resolution (CSR) is proposed as an advanced task, which greatly meets the requirements of industry and human life. In general, the low-quality images in CSR are jointly degraded with compression artifacts, \eg, block artifacts, ring effects, and low-resolution artifacts. The severe and heterogeneous degradation poses more challenges and high requirements for the CSR backbones. Moreover, in real applications, the compression codecs are usually diverse in different platforms, which urgently entails the Universal CSR model.

There are some pioneering works~\cite{li2023cswin2sr,conde2022swin2sr,li2022hst,stablesr} attempting to remove this hard degradation by improving the representation ability. The representative strategy is to design the CSR backbone with the Transformer, which profits from the self-attention module. For instance, Swin2SR~\cite{conde2022swin2sr} introduces the enhanced Swin Transformer~\cite{liu2021swin,liu2022swin} (\ie, SwinV2) to boost the restoration capability of the CSR backbone. HST~\cite{li2022hst} utilizes the hierarchical backbone to excavate multi-scale representation for CSR.  Despite the transformer-based backbones having revealed strong recovery capability in CSR, the high computational cost of the transformer prevents its application and training optimization~\cite{liang2021swinir,conde2022swin2sr}. 
Recently,  Multi-layer perceptron (MLP) has demonstrated its potential to achieve the trade-off between the computational cost and global dependency modeling in the  classification~\cite{chen2021cyclemlp,lian2021mlpAsMLP,wei2022activemlp,touvron2022resmlpResMLP,tolstikhin2021mlpMLP-Mixer}, benefiting from its efficient and effective token mixer strategies. Inspired by this, the first MLP-based framework MAXIM~\cite{tu2022maxim} in image processing is proposed,  where the image tokens interact in global and local manners with multi-axis MLP, respectively. However, the above works only focus on single distortion removal, which lacks enough universality for CSR tasks. 

In this work, we propose the first universal framework, dubbed UCIP, for CSR tasks with our dynamic prompt strategy based on an MLP-like module. It is noteworthy that the optimal contextual information obtained with the CSR network tends to vary with the content/spatial and degradation type, which entails the content-aware task-adaptive contextual information modeling capability. To achieve this, existing prompt-based IR~\cite{promptir,PIP,li2023diffusion,li2024promptcir} methods have attempted to set multiple prompts with image size, lacking adaptability for various input sizes and leading to more computational cost. In contrast, our dynamic prompt strategy can not only achieve content-aware task-adaptive modulation but also own more applicability. Concretely, we propose the Dynamic Prompt generation Module (DPM), where a group of prompts with the size of $1\times1\times C_p$ is set and $C_p$ is the channel dimension. Then spatial-wise composable coefficients $H \times W \times C_p$ are generated with the distorted images, which guides the cooperation of these prompt bases to form the dynamic prompt with image size, thereby owning the content/spatial- and task-adaptive modulation capability.

Based on the powerful DPM, we can achieve the universal CSR framework by incorporating it into existing CSR backbones. However, in the commonly used Transformer backbone, contextual information modeling is achieved with the cost attention module, where any two tokens are required to interact. In contrast, an active token mixer (ATM)~\cite{wei2022activemlp} has been proposed for the MLP-like backbone to reduce the computational cost by implicitly achieving contextual information modeling in the horizontal and vertical directions with offset generation. However, no works explore the potential of this backbone on low-level vision tasks. Inspired by this, we propose the dynamic prompt-guided token mixer block (PTMB) by fusing the advantages of our DPM and ATM, where our DPM can guide the contextual information modeling process of the ATM by modulating the offset prediction and toke mixer. Notably, only horizontal and vertical contextual modeling in ATMs lacks enough local information utilization. Consequently, we increase a local branch in PTMB with one $3 \times 3$ convolution. Based on PTMB, our UCIP can achieve efficient and excellent universal compressed image super-resolution for different codecs/modes.

To build the benchmark dataset for universal CSR tasks, we collected the datasets with 6 representative image codecs, including 3 traditional codecs and 3 learning-based codecs. Concretely, traditional codecs consist of JPEG~\cite{JPEG}, all-intra mode of HEVC~\cite{HM}, and VVC~\cite{VTM}. For learning-based codecs, to ensure the diversity of degradations, we select 3 codecs with different optimization objectives, \ie, PSNR-oriented, SSIM-oriented, and GAN-based codecs. In this way, our database can cover the prominent compression types in recent industry and research fields. We have compared our UCIP and reproduced state-of-the-art methods on this benchmark, which showcases the superiority and robustness of our UCIP.

The contributions of this paper are listed as follows:
\begin{itemize}
    \item We propose the first universal framework, \ie, UCIP for the CSR tasks with our dynamic prompt strategy, intending to achieve the "all-in-one" for the CSR degradations with different codecs/modes.  
    \item We propose the dynamic prompt-guided token mixer block (PTMB) by fusing the advantages of our proposed dynamic prompt generation module (DPM) and revised active token mixer (ATM), as the basic block for UCIP. 
    \item We propose the first dataset benchmark for universal CSR tasks by collecting datasets with 6 prominent traditional and learning-based codecs, consisting of multiple compression degrees. This ensures the diversity of degradations in the benchmark dataset, thereby being reliable as the benchmark to measure different CSR methods. 
    \item Extensive experiments on our universal CSR benchmark dataset have revealed the effectiveness of our proposed UCIP, which outperforms the recent state-of-the-art transformer-based methods with lower computational costs. 
\end{itemize}

\section{Related Works}

\subsection{Compressed Image Super-resolution}
Compressed Image Super-resolution aims to tackle complicated hybrid distortions, including compression artifacts and low-resolution artifacts~\cite{fbcnn,yang2023aim2022,li2022hst,li2023cswin2sr,li2021task,csr5,wu2021learned,IPT}. The first challenge for this task was held in the AIM2022~\cite{yang2023aim2022}, where the image is first downsampled with the bicubic operation and then compressed with a JPEG codec. To solve this hard degradation, some works~\cite{li2023cswin2sr,conde2022swin2sr,qin2023cidbnetHAT,li2022hst} seek to utilize the Transformer-based architecture as their backbone. For instance, Swin2SR~\cite{conde2022swin2sr} eliminates the training instability and the requirements for large data for CSR by incorporating the Swin Transformer V2 to SwinIR~\cite{liang2021swinir}. HST~\cite{li2022hst} utilizes the multi-scale information flow and pre-training strategy~\cite{li2021few} to enhance the restoration process with a hierarchical swin transformer. To further fuse the advantages of convolution and transformer, Qin \etal \cite{qin2023cidbnetHAT} proposes a dual-branch network, which achieves the consecutive interaction between the convolution branch and transformer branch. In contrast, to achieve the trade-off between the performance and computational cost, we aim to explore one efficient and effective framework for universal CSR problem. 

\subsection{MLP-like Models}
\label{mlp-like}
As the alternative model for Transformer and Convolution Neural Networks (CNNs), MLP-like models~\cite{lian2021mlpAsMLP,chen2021cyclemlp,tolstikhin2021mlpMLP-Mixer,wei2022activemlp,zhang2021morphmlp,touvron2022resmlpResMLP,tang2022imageWaveMLP,tang2022sparsesMLP,hou2022visionViP,yu2022s2S2MLP} have attracted great attention for their concise architectures. Typically, the noticeable success of MLP-like models stems from the well-designed token-mixing strategies~\cite{wei2022activemlp}. The pioneering works, MLP-Mixer~\cite{tolstikhin2021mlpMLP-Mixer} and ResMLP~\cite{touvron2022resmlpResMLP} adopt two types of MLP layers, \ie, channel-mixing MLP and token-mixing MLP, which are responsible for the channel and spatial information interaction. To simplify the token-mixing MLP, Hou \etal~\cite{hou2022visionViP} and Tang \etal~\cite{tang2022sparsesMLP} decompose the token-mixing MLP into the horizontal and vertical token-mixing MLPs. Sequentially,  As-MLP~\cite{lian2021mlpAsMLP} introduces the two-axis token shift in different channels to achieve global token mixing. There are also several works that take the hand-craft windows to enlarge the receptive field for better spatial token mixing, \eg, WaveMLP~\cite{tang2022imageWaveMLP}, and MorphMLP~\cite{zhang2021morphmlp}. However, the token-mixing strategies in the above methods are restrictively fixed and lack flexibility and adaptability for different contents. To overcome this, ATM~\cite{wei2022activemlp} is proposed to achieve the active token selection and mixing in each channel. Based on the progress of the above MLP-like models, MAXIM~\cite{tu2022maxim} is the first work to introduce the MLP-like model in low-level processing. However, the potential of MLP-like models is yet to be explored, as restoration model not only requires long-range token mixing but also demands efficient local feature extractions.

\subsection{Prompt Learning}

In the field of Natural Language Processing (NLP), prompt learning has emerged as a pivotal technique, particularly with the advent of transformer-based pre-trained models such as GPT~\cite{GPT3,openai2023gpt4} and BERT~\cite{devlin2018bert}. Prompt learning involves providing models with specific textual cues that guide their processing of subsequent input, which helps models fast adapt to unseen tasks or applications. This approach has proven instrumental in directing models for task-specific outputs without necessitating extensive retraining or fine-tuning. Despite the success in NLP tasks, some researchers adopt prompt learning into vision tasks~\cite{visual_prompt_tuning,visual_prompt_tuning2,li2021prefix,ai2023multimodal,prompt3,prompt4,wang2024promptrr}. Among them, PromptIR~\cite{promptir} is the first to explore the low-level restoration model with prompts to facilitate multi-task learning~\cite{vandenhende2022multi,vandenhende2020mti,Airnet,su2015multi,gao2023prompt}. Prompts here act as a small set of learnable parameters which interact with image features during training, providing task-specific guidance. Therefore, the prompts should be as much dynamic as possible to adapt to various degradation tasks and different pixel distributions.
\begin{figure}[t]
    \centering
    \includegraphics[width=0.98\linewidth]{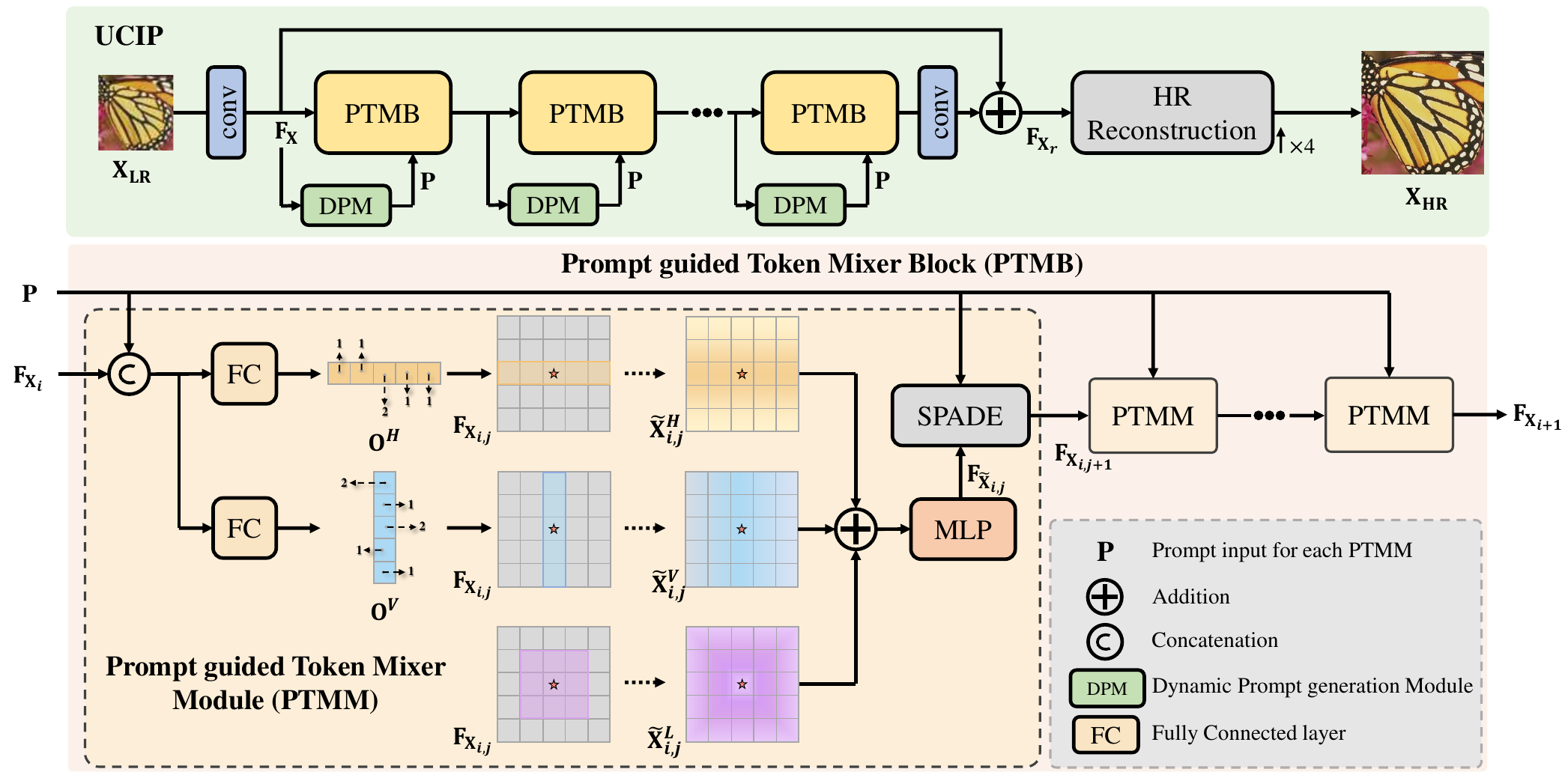}
    \caption{Illustration of our proposed UCIP. From top to bottom: (a) The overall framework of UCIP. The LR is first enhanced through several consecutive PTMBs, then upsampled by HR reconstruction module. (b) The architecture of PTMB. Each PTMB utilizes the dynamic prompt generated from a DPM and several cascading PTMMs to iteratively refine distorted inputs. (c) The architecture of PTMM. PTMM takes prompt \textbf{P} along with image feature $\textbf{F}_{{\text{X}}_i}$ as input to adaptively generate offsets, which facilitate the network to perform content/spatial-aware task-adaptive contextual information extraction.}
    \label{fig:framework}
\end{figure}

\section{Methods}
In this section, we first clarify the principle and construction of our dynamic prompt generation module in Sec.~\ref{sec:dpm}, and then describe how to achieve the basic block of our UCIP, \ie, dynamic prompt-guided token mixer block in Sec.~\ref{sec:dpmb}. Finally, we depict the whole framework of our UCIP in Sec.~\ref{sec:op}. 
\subsection{Dynamic Prompt Generation Module}
\label{sec:dpm}

As stated in Sec.~\ref{sec:intro}, the universal CSR tasks entail the content/spatial- and task-adaptive modulation. An intuitive strategy is to set one prompt with the image size for each task individually or fuse them adaptively. However, it will bring severe parameter costs with the increase of the task number or image size~\cite{promptir}. To mitigate this, we propose the dynamic prompt strategy, and design the corresponding dynamic prompt generation module (DPM), intending to only exploit a small amount of prompt with $1\times1\times C_p$ and achieve the content/spatial- and task-adaptive with the cooperation of them. To this end, we decouple the large dynamic prompt with the size of $H\times W \times C_p$ into two smaller matrices, \ie, the coefficients $\mathbf{w_I}$ with the size of $H\times W \times D$ and $D$ basic prompts with the size of $1 \times 1 \times C_p$. We can understand that for each spatial position $\{i, j\}$, there is one group of coefficients $w_I(i, j)$ to combine $D$ basic prompts. thereby being content/spatial-adaptive. To let the dynamic prompt perceive the task information, we generate the coefficients with the feature of input images directly, thereby being task-adaptive and suitable for any input size.  Our implementation has two advantages: 1) no extra operations to adjust the spatial size of prompts, and thus the guidance information from prompts is explicit and accurate; 2) our prompts have fewer parameters and are more computationally-friendly compared to previous methods~\cite{promptir}.

\begin{wrapfigure}{r}{0.5\textwidth}
    \centering
    \includegraphics[width=0.48\textwidth]{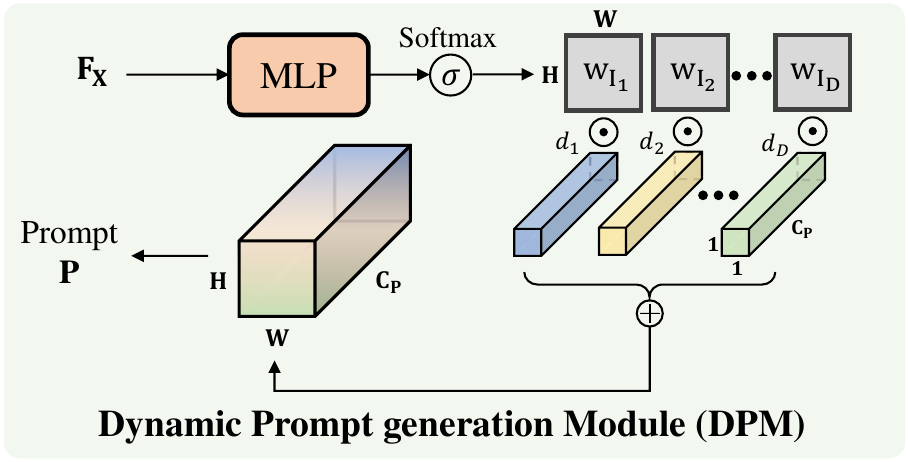}
    \caption{The architecture of DPM. To dynamically aggregate content/spatial-aware task-adaptive contextual information, we introduce few number of basic dynamic kernels into the generation process of our prompt. Moreover, our design maintains adaptability to arbitrary input resolutions.}
    \label{fig:DPM}
\end{wrapfigure}

The overall architecture of $\operatorname{DPM}$ is shown in Fig.~\ref{fig:DPM}, where the learnable basic prompts $\textbf{P}_{\text{I}} \in \mathbb{R}^{D \times 1 \times 1 \times C_{P}}$ are set. Here, the $D$ and $C_P$ are the number of base prompts and the channel dimension of prompts. To generate dynamic prompt coefficients from input features $\textbf{F}_{\text{X}} \in \mathbb{R}^{H \times W \times C}$, an MLP layer is applied to extract the degradation prior and transform the channel dimension from $C$ to the number of basic prompts $D$. Then, the $\operatorname{softmax}$ operation is exploited to generate the composable coefficients $\textbf{w}_{\text{I}} \in \mathbb{R}^{D \times H \times W \times 1}$ for basic prompts. Based on the inversion of the above dynamic prompt decomposition, we can obtain the dynamic prompt as:
\begin{align}
     \textbf{w}_{\text{I}} = \operatorname{Softmax}(\operatorname{MLP}(\textbf{F}_{\text{X}})),\quad
     \textbf{P} = \sum^{D}\left(\textbf{w}_{\text{I}} \odot \textbf{P}_{\text{I}}\right)
\end{align}

\subsection{Prompt-guided Token Mixer Block}
\label{sec:}

\subsubsection{Prompt-guided token mixer module}
\label{sec:dpmb}
After obtaining the dynamic prompt, we can exploit it to guide the restoration network for universal CSR tasks. Recently, Active Token Mixer (ATM)~\cite{wei2022activemlp} gain great success in high-level vision tasks due to their well-designed token-mixing strategies. In contrast to transformer architecture, where the contextual information modeling is performed with the interactions between any two tokens, ATM utilize the deformable convolution to predict the offset of mostly relevant tokens, achieving the implicit contextual
information modeling in the horizontal and vertical directions with offset generation.

Inspired by this, we propose the Dynamic Prompt-guided Token Mixer Module, dubbed PTMM by exploiting the dynamic prompt generated with DPM to guide the prediction of the offset of most informative tokens for contextual modeling. Concretely, PTMM leverages deformable convolutions and offsets to adaptively fuse tokens across horizontal and vertical axes, regardless of diverse degradation. However, as mentioned in~\cite{tu2022maxim}, MLP-like modules exhibit diminished efficacy in the extraction of local relevance, which is essential for compressed super-resolution tasks. Therefore, we introduce a depth convolution around the target pixel to achieve the local information extraction.

As shown in Fig.~\ref{fig:framework}(b), PTMM first extracts vertical and horizontal representative offsets $\mathbf{O}^V, \mathbf{O}^H$ by two sets of fully connected layers. To incorporate task-adaptive information during offset generation, we concatenate dynamic prompt generated from DPM with input features $\textbf{F}_{\text{X}}$ as the condition:
\begin{equation}
    \mathbf{O}^{\{V,H\}} = \operatorname{FC}_{\{V,H\}}(\operatorname{Concat}([\textbf{F}_{\text{X}}, \textbf{P}]))
\end{equation}
Then, we use the offset to recompose features along one certain axis into a new token $\tilde{\mathbf{x}}^{\{V,H\}}$ by the deformable convolution for information fusion (\ie, token mixer). 
In addition, we adopt a depth convolution to achieve the local information extraction: 
\begin{equation}
    \tilde{\mathbf{x}}^{L} = \operatorname{Conv_{3\times 3}}(\textbf{F}_{\text{X}})
\end{equation}
After we obtain these three tokens $\tilde{\mathbf{x}}^{\{V,H,L\}}$, we adaptively mix them with learned weights, formulated as
\begin{equation}
\textbf{F}_{\tilde{\mathbf{x}}}=\boldsymbol{\alpha}^V \odot \tilde{\mathbf{x}}^V+\boldsymbol{\alpha}^H \odot \tilde{\mathbf{x}}^H+\boldsymbol{\alpha}^L \odot \tilde{\mathbf{x}}^L
\end{equation}
where $\odot$ denotes element-wise multiplication. $\boldsymbol{\alpha}^{\{V, H, L\}} \in$ $\mathbb{R}^C$ are learned from the summation $\tilde{\mathbf{x}}^{\Sigma}$ of $\tilde{\mathbf{x}}^{\{V, H, L\}}$ with weights $W^{\{V, H, L\}} \in \mathbb{R}^{C \times C}$, where $C$ denotes the channel dimension:
$$
\left[\boldsymbol{\alpha}^V, \boldsymbol{\alpha}^H, \boldsymbol{\alpha}^L\right]=\sigma\left(\left[W^V \cdot \tilde{\mathbf{x}}^{\Sigma}, W^H \cdot \tilde{\mathbf{x}}^{\Sigma}, W^L \cdot \tilde{\mathbf{x}}^{\Sigma}\right]\right),
$$
Here, $\sigma(\cdot)$ is a softmax function for normalizing each channel separately.

To further incorporate the task prior for our UCIP, we modulate mixed features $\textbf{F}_{\tilde{\mathbf{x}}}$ using the aforementioned dynamic prompt $\textbf{P}$ by a SPADE block~\cite{SPADE} as the output features of the PTMM, which is shown in the  Fig.~\ref{fig:framework}.

\subsubsection{Discussions}
There are two most relevant MLP-like methods, \ie, MAXIM~\cite{tu2022maxim} and ActiveMLP~\cite{wei2022activemlp}. The differences between MAXIM and our UCIP are as: MAXIM is only designed for specific task, where the cross-gating block and dense connection result in severe computational costs. The differences between ActiveMLP and our UCIP are as: ActiveMLP is designed for classification and focuses more on global information extraction, lacking local perception. Compared with them, our UCIP introduces the simple MLP-based architecture and the dynamic prompt for low-level vision, which is more applicable than the above methods for Universal CSR.

\subsubsection{Overall pipeline}
To improve the modeling cability of PTMB, we connect $N$ PTMMs in a successive way. It is worth noting that, to balance the performance of model and the computational cost, we share the prompt \textbf{P} across all PTMMs within a single PTMB. With respect to offsets, we generate new offsets every two PTMMs. The whole process of PTMB can be formulated as:
\begin{equation}
    \textbf{P} = \operatorname{DPM}(\textbf{F}_{\text{X}}, \textbf{P}_{\text{I}}),\quad
    \textbf{F}_{\text{X}_{i+1}} = \operatorname{PTMM}(\textbf{P}, \textbf{F}_{\text{X}_{i}})
\end{equation}
where $\textbf{F}_{\text{X}_{i}}$ is the input feature of $i^{th}$ PTMM.

\subsection{Overall Framework}
\label{sec:op}
As shown in Fig.~\ref{fig:framework}, we build our UCIP following the popular pipeline of compressed super-resolution backbones, which is composed of shallow feature extraction, deep feature restoration, and HR reconstruction modules. Given a low-resolution input image $\textbf{X}_{\text{LR}} \in \mathbb{R}^{H \times W \times 3}$,  UCIP first extracts the shallow features $\textbf{F}_{\text{X}} \in \mathbb{R}^{H \times W \times C}$ using a patch-embedding layer, where $H$, $W$ are the spatial dimensions of features. Then, we pass $\textbf{F}_{\text{X}}$ through several PTMB to recursively remove the compression artifacts and generate the restored features $\textbf{F}_{\text{X}_{r}}$. Finally, following~\cite{wang2018esrganRRDB,liang2021swinir}, we use a series of convolution layers and nearest interpolation operations to obtain the final high-resolution output $\textbf{X}_{\text{HR}}$, which can be represented as:
\begin{equation}
    \textbf{X}_{\text{HR}} =\operatorname{Conv}(\operatorname{Conv}(\operatorname{Conv}(\textbf{F}_{\text{X}} + \textbf{F}_{\text{X}_{r}})\uparrow_{\times 2})\uparrow_{\times 2})
\end{equation}

\subsection{Our UCSR Dataset}
\label{general}
To facilitate current and future research in CSR, we propose the first benchmark dataset for universal CSR, dubbed UCSR dataset, which not only considers traditional compression methods but also learning-based compression methods. We consider 6 types of compression codecs, including 3 most representative traditional codecs JPEG~\cite{JPEG}, HM~\cite{HM}, VTM~\cite{VTM}, and 3 open-sourced learning-based codecs $\text{Cheng}_\text{PSNR}$~\cite{cheng2020learned}, $\text{Cheng}_\text{SSIM}$~\cite{cheng2020learned} (abbreviated as $\text{C}_\text{PSNR}$ and $\text{C}_\text{SSIM}$ in the following paper), HIFIC~\cite{mentzer2020hifi}. Thesse three learning-based codecs are PSNR-oriented and SSIM-oriented variants from~\cite{cheng2020learned} and perceptual-oriented GAN-based codecs from~\cite{mentzer2020hifi}, respectively. To cover the prominent compression types in real scenarios, we consider four different compression qualities for each codec, except for HIFIC, since only the weights for three bitrate points are released. 

To generate the training dataset, we choose the popular DF2K~\cite{DIV2K,Flickr2K}, which contains 3450 high-quality images. Each image is downsampled by a scale factor of 4 using MATLAB bicubic algorithm. Then, we compress the downsampled images with six different compression algorithms to yield the training dataset of all competitive methods and our UCIP. The quality factors we used for different codecs are respectively as: (i) [10, 20, 30, 40] for JPEG, where the smaller value means poorer image quality. (ii) [32, 37, 42, 47] for HM, VTM, where value denotes the quantization parameter (QP), and larger value means poor quality. (iii) [1, 2, 3, 4] for $\text{C}_\text{PSNR}$, $\text{C}_\text{SSIM}$, where the smaller value indicates poorer quality. We adopt the implementation in the popular open-sourced compression tools compressai~\cite{begaint2020compressai}. (iv) [`low', `med', `high'] for HIFIC, where `low' indicates the poorest image quality. We use the PyTorch implementation~\cite{hific_pytorch} to compress images. All the methods are trained from scratch on our proposed benchmarks. We adopt the same process to generate the evaluation datasets based on five commonly used benchmarks: Set5~\cite{Set5}, Set14~\cite{Set14}, BSD100~\cite{BSDall}, Urban100~\cite{urban100} and Manga109~\cite{manga109}.

\section{Experiments}

\begin{table*}[htp]
\centering
\caption{Quantitative comparison for compressed image super-resolution on traditional codecs. Results are tested on $\times4$ with different compression qualities in terms of  PSNR$\uparrow$/SSIM$\uparrow$. The best performances are in \tcr{red}. \textit{Notably, all compared methods are trained from scratch with our proposed UCSR dataset for fair comparisons.} $\mathcal{D}$ denotes for ``Datasets''.}
\setlength{\tabcolsep}{2pt}
\resizebox{\textwidth}{!}{
\begin{tabular}{c|c|cccc|cccc|cccc}
\hline
\multirow{2}{*}{$\mathcal{D}$} & \multirow{2}{*}{Methods} & \multicolumn{4}{c|}{JPEG~\cite{JPEG}}                             & \multicolumn{4}{c|}{HM~\cite{HM}}                               & \multicolumn{4}{c}{VTM~\cite{VTM}}                               \\ \cline{3-14} 
                          &                          & $\mathcal{Q}=10 $         & $\mathcal{Q}=20 $          & $\mathcal{Q}=30 $          & $\mathcal{Q}=40 $         & $\mathcal{Q}=47 $          & $\mathcal{Q}=42 $          & $\mathcal{Q}=37 $          & $\mathcal{Q}=32 $          & $\mathcal{Q}=47 $          & $\mathcal{Q}=42 $          & $\mathcal{Q}=37 $          & $\mathcal{Q}=32 $         \\ \hline
\multirow{7}{*}{\rotatebox{90}{Set5~\cite{Set5}}}     & RRDB~\cite{wang2018esrganRRDB}                     & 24.44/0.676 & 25.93/0.729 & 26.70/0.754 & 27.22/0.769 & 22.48/0.624 & 24.48/0.690 & 26.52/0.752 & 28.05/0.794 & 22.70/0.635 & 24.84/0.706 & 26.65/0.758 & 28.10/0.797 \\
                          & SwinIR~\cite{liang2021swinir}                   &        24.79/0.703     &      26.25/0.747       &    27.07/0.771         &       27.59/0.783      &       22.66/0.647      &       24.54/0.703      &       26.82/0.765      &        28.53/0.809     &     22.81/0.652        &     24.97/0.716        &     26.93/0.768        &       28.72/0.813      \\
                          & Swin2SR~\cite{conde2022swin2sr}                  &       24.80/0.705      &     26.24/0.752        &    27.16/0.774         &     27.64/0.786        &      22.66/0.650       &      24.55/0.705       &    26.81/0.766         &       28.50/0.809      &        22.79/0.652     &       24.91/0.716      &     26.89/0.769        &      28.64/0.813       \\
                          & MAXIM~\cite{tu2022maxim}                    &          24.83/0.709   &      26.15/0.751       &     27.00/0.773        &       27.44/0.784      &       22.69/0.648      &      24.60/0.705       &      26.75/0.764       &      28.48/0.808       &        22.88/0.654     &       24.96/0.718      &      26.89/0.770       &        28.61/0.811     \\
                          & AIRNet~\cite{Airnet}                   &      24.67/0.701       &      26.04/0.745       &      26.83/0.767       &      27.30/0.779       &        22.56/0.640     &     24.38/0.698        &      26.55/0.760      &       28.24/0.805      &      22.71/0.648       &     24.81/0.714        &      26.65/0.766       &        28.38/0.810     \\
                          & PromptIR~\cite{promptir}                 &     24.82/0.707        &       26.24/0.751      &      27.13/0.774       &      27.62/0.787       &        22.68/0.652     &    24.55/0.705         &      26.87/0.768       &       28.64/0.813      &       \tcr{22.89/0.658}      &       24.99/0.720      &        26.93/0.771     &        28.74/0.815     \\
                          & UCIP                &        \tcr{25.05/0.715}    &     \tcr{26.53/0.761}      &     \tcr{27.44/0.782}      &   \tcr{27.94/0.794}       &        \tcr{22.77/0.656}    &      \tcr{24.76/0.711}     &      \tcr{27.05/0.772}      &       \tcr{28.82/0.815}     &       \tcr{22.89}/0.657      &        \tcr{25.11/0.722}     &      \tcr{27.17/0.775}     &    \tcr{28.95/0.819}        \\ \hline\hline
\multirow{7}{*}{\rotatebox{90}{Set14~\cite{Set14}}}    & RRDB~\cite{wang2018esrganRRDB}                     &      23.40/0.579       &       24.49/0.619      &       25.01/0.639      &      25.32/0.651       &      21.84/0.531       &      23.48/0.584       &       24.93/0.635      &       25.99/0.679      &    22.12/0.541         &      23.74/0.594       &      25.09/0.643       &      26.05/0.682       \\
                          & SwinIR~\cite{liang2021swinir}                   &        23.77/0.596     &     24.81/0.630        &    25.32/0.649         &        25.66/0.662     &     21.96/0.542        &      23.59/0.593       &     25.13/0.645        &       26.38/0.695      &     22.19/0.550        &     23.79/0.599        &      25.32/0.652       &       26.49/0.699      \\
                          & Swin2SR~\cite{conde2022swin2sr}                  &       23.79/0.597     &     24.84/0.631        &     25.36/0.651        &      25.68/0.663       &      21.97/0.543       &        23.59/0.594     &     25.17/0.646        &      26.42/0.697       &         22.18/0.550    &     23.77/0.600        &       25.30/0.652&       26.48/0.700      \\
                          & MAXIM~\cite{tu2022maxim}                    &          23.79/0.597   &     24.83/0.632        &      25.33/0.651       &       25.66/0.663      &       22.02/0.543      &       23.60/0.593      &      25.15/0.645       &     26.39/0.694        &       22.24/0.551      &       23.83/0.601      &      25.33/0.653       &       26.48/0.699      \\
                          & AIRNet~\cite{Airnet}                   &      23.61/0.593       &      24.64/0.629       &      25.13/0.647       &       25.43/0.659      &       21.90/0.540      &     23.47/0.591        &        24.97/0.642     &      26.18/0.691       &       22.11/0.548      &       23.68/0.598     &       25.12/0.650      &        26.24/0.696     \\
                          & PromptIR~\cite{promptir}                 &    23.79/0.599         &      24.84/0.634       &    25.34/0.652         &        25.67/0.664     &        21.99/0.544     &      23.53/0.594       &       25.17/0.647      &       26.44/0.697      &      22.21/0.552       &      23.78/0.601       &      25.34/0.654       &     26.50/0.701        \\
                          & UCIP                &     \tcr{23.93/0.602}        &      \tcr{24.99/0.637}     &       \tcr{25.53/0.657}      &    \tcr{25.88/0.669}         &      \tcr{22.10/0.547}      &       \tcr{23.70/0.597}      &        \tcr{25.34/0.650}     &     \tcr{26.63/0.701}        &       \tcr{22.28/0.553}      &    \tcr{23.89/0.603}         &     \tcr{25.45/0.656}        &     \tcr{26.71/0.705}        \\ \hline\hline
\multirow{7}{*}{\rotatebox{90}{BSD100~\cite{BSDall}}}    & RRDB~\cite{wang2018esrganRRDB}                     &     23.56/0.547        &      24.44/0.580       &     24.86/0.597        &     25.12/0.609        &       22.10/0.503      &      23.43/0.542       &     24.64/0.588        &      25.58/0.630       &     22.30/0.510        &       23.64/0.550      &     24.80/0.595        &     25.66/0.634        \\
                          & SwinIR~\cite{liang2021swinir}                   &        23.79/0.557     &     24.62/0.587        &    25.04/0.604         &       25.31/0.616      &       22.17/0.510      &       23.45/0.548      &      24.74/0.596       &    25.80/0.643         &       22.34/0.516      &     23.66/0.555        &      24.91/0.603       &      25.92/0.649       \\
                          & Swin2SR~\cite{conde2022swin2sr}                  &       23.79/0.557      &      24.62/0.588       &   25.03/0.605          &      25.30/0.617       &  22.15/0.511           &      23.42/0.549       &    24.72/0.596         &      25.81/0.645       &    22.32/0.516         &      23.60/0.555       &      24.88/0.603       &     25.91/0.650        \\
                          & MAXIM~\cite{tu2022maxim}                    &          23.81/0.558   &      24.63/0.589       &     25.04/0.606        &      25.30/0.618       &      22.20/0.510       &       23.49/0.548      &       24.73/0.595      &     25.79/0.644        &     22.36/0.516        &       23.67/0.556      &      24.89/0.603       &       25.90/0.650      \\
                          & AIRNet~\cite{Airnet}                   &     23.73/0.555        &      24.55/0.586       &      24.95/0.603       &      25.22/0.615       &      22.13/0.509       &      23.42/0.547       &       24.68/0.594      &     25.71/0.642        &       22.31/0.515      &      23.62/0.554       &       24.83/0.602      &        25.81/0.648     \\
                          & PromptIR~\cite{promptir}                 &     23.82/0.559        &       24.65/0.589      &     25.05/0.606        &       25.32/0.618      &         22.20/0.511    &     23.48/0.549        &      24.75/0.597       &        25.82/0.645     &       22.35/\tcr{0.517}      &       23.66/0.556      &        24.91/0.604     &       25.93/0.651      \\
                          & UCIP                &       \tcr{23.88/0.561}      &  \tcr{24.73/0.593}           &       \tcr{25.15/0.610}      &     \tcr{25.42/0.623}        &      \tcr{22.24/0.513}       &     \tcr{23.56/0.551}     &       \tcr{24.84/0.599}     &     \tcr{25.93/0.649}        &      \tcr{22.38/0.517}       &     \tcr{23.74/0.558}        &      \tcr{24.99/0.606}       &      \tcr{26.03/0.654}      \\ \hline\hline
\multirow{7}{*}{\rotatebox{90}{Urban100~\cite{urban100}}}    & RRDB~\cite{wang2018esrganRRDB}                     &       21.69/0.578      &      22.18/0.597       &      22.66/0.622       &       22.97/0.638      &     20.42/0.531        &     21.66/0.578        &       22.84/0.633      &        23.61/0.671     &     20.67/0.543        &       21.95/0.593      &     23.00/0.641        &       23.66/0.674      \\
                          & SwinIR~\cite{liang2021swinir}                   &         21.74/0.580    &       22.61/0.621      &    23.11/0.646         &      23.41/0.661       &      20.45/0.535       &       21.86/0.595      &       23.18/0.654      &       24.12/0.699      &        20.70/0.546     &      22.10/0.607       &      23.33/0.662       &       24.19/0.703      \\
                          & Swin2SR~\cite{conde2022swin2sr}                  &       21.79/0.582      &     22.67/0.624        &     23.17/0.648        &      23.44/0.664       &    20.48/0.536         &       21.90/0.597      &    23.21/0.655         &       24.16/0.700      &    20.72/0.548         &     22.11/0.608        &           23.34/0.662  &       24.22/0.703      \\
                          & MAXIM~\cite{tu2022maxim}                    &          21.78/0.582   &    22.61/0.622         &     23.08/0.645        &      23.38/0.660       &     20.47/0.534        &    21.87/0.594         &      23.13/0.651       &       24.05/0.695      &      20.72/0.547       &      22.10/0.606       &       23.28/0.659      &       24.11/0.698      \\
                          & AIRNet~\cite{Airnet}                   &        21.57/0.574     &       22.40/0.615      &       22.86/0.639      &       23.14/0.655      &      20.35/0.530       &       21.72/0.590      &      22.97/0.648       &       23.87/0.692      &      20.60/0.543       &      21.96/0.603       &     23.12/0.657        &        23.92/0.696     \\
                          & PromptIR~\cite{promptir}                 &     21.81/0.587        &      22.65/0.626       &     23.12/0.649        &     23.42/0.664        &      20.50/0.539       &     21.89/0.598        &     23.17/0.656        &      24.13/0.701       &     20.73/0.550        &       22.11/0.609      &       23.32/0.663      &      24.18/0.704       \\
                          & UCIP                &      \tcr{22.00/0.596}       &      \tcr{22.88/0.637}      &       \tcr{23.39/0.664}      &    \tcr{23.71/0.677}        &      \tcr{20.59/0.542}      &        \tcr{22.05/0.604}    &      \tcr{23.39/0.661}     &    \tcr{24.42/0.711}       &      \tcr{20.80/0.552}       &   \tcr{22.23/0.614}         &      \tcr{23.50/0.670}     &      \tcr{24.46/0.715}      \\ \hline\hline
\multirow{7}{*}{\rotatebox{90}{Manga109~\cite{manga109}}}    & RRDB~\cite{wang2018esrganRRDB}                     &        22.50/0.684     &     23.75/0.730        &       24.49/0.756      &        24.99/0.773     &       21.17/0.655      &       23.24/0.722      &       25.07/0.778      &        26.24/0.813     &      21.59/0.675       &       23.64/0.738      &      25.27/0.786       &         26.29/0.815    \\
                          & SwinIR~\cite{liang2021swinir}                   &        23.05/0.720     &      24.38/0.762       &    25.16/0.786         &        25.67/0.801     &     21.40/0.677        &     23.56/0.743        &        25.64/0.801     &        27.17/0.841     &      21.73/0.689       &          23.90/0.754   &        25.83/0.807     &      27.25/0.843       \\
                          & Swin2SR~\cite{conde2022swin2sr}                  &        23.09/0.720     &      24.40/0.762       &    25.18/0.786         &     25.69/0.801        &   21.42/0.677          &      23.58/0.743       &     25.62/0.799        &      27.11/0.839       &    21.75/0.690         &      23.90/0.753       &            25.78/0.804 &     27.19/0.841        \\
                          & MAXIM~\cite{tu2022maxim}                    &          23.11/0.722   &      24.41/0.762       &    25.17/0.786         &       25.65/0.800      &        21.41/0.675     &    23.55/0.740         &     25.56/0.797        &        27.05/0.836     &       21.74/0.688      &      23.89/0.752       &      25.76/0.803       &       27.13/0.838      \\
                          & AIRNet~\cite{Airnet}                   &      22.82/0.714       &       24.07/0.754      &     24.78/0.778        &      25.26/0.793       &      21.25/0.670       &      23.34/0.735       &      25.29/0.793       &      26.69/0.833       &        21.59/0.684     &      23.67/0.747       &       25.47/0.800      &       26.74/0.835      \\
                          & PromptIR~\cite{promptir}                 &     23.15/0.726        &      24.48/0.767       &     25.23/0.789        &      25.71/0.804       &        21.41/0.681     &    23.59/0.746         &     25.62/0.801        &       27.15/0.841      &      21.73/0.692       &       23.90/0.755      &       25.80/0.807      &        27.21/0.843     \\
                          & UCIP                &      \tcr{23.36/0.734}      &       \tcr{24.77/0.775}     &      \tcr{25.58/0.798}       &     \tcr{26.11/0.813}       &     \tcr{21.54/0.683}       &     \tcr{23.79/0.750}       &       \tcr{25.94/0.808}     &    \tcr{27.61/0.848}       &     \tcr{21.82/0.693}        &       \tcr{24.06/0.759}&       \tcr{26.08/0.812}      &      \tcr{27.68/0.850}       \\ \hline
\end{tabular}
}
\label{table:traditional}
\end{table*}

Our objective is to develop an MLP-like model that caters to a wide range of compressed image super-resolution tasks. Thus, we evaluate our UCIP on six different CSR tasks, including three traditional compression codecs: JPEG~\cite{JPEG}, HM~\cite{HM}, VTM~\cite{VTM}; and three learning-based compression codecs: $\text{C}_\text{PSNR}$~\cite{cheng2020learned}, $\text{C}_\text{SSIM}$~\cite{cheng2020learned}, HIFIC~\cite{mentzer2020hifi}. 

\subsubsection{Implement details} We train our UCIP from scratch in an end-to-end manner. We employ an Adam optimizer with initial learning rate of 3e-4. The learning rate is halved after 200k iterations, and the total number of iterations is set to 40w. The network is optimized by L1 loss. During training, we randomly cropped degraded low-resolution images into patches of size $64 \times 64$, and $256 \times 256$ for high-resolution counterparts as well. Following previous works, random horizontal and vertical flips are utilized to augment training data. The total batch size is set to 32. For our baseline model, we use 6 PTMBs for UCIP and 6 PTMMs for each PTMB.

\subsubsection{Training details} To ensure fair comparisons, we train all the competitive methods following their official released codes on our proposed CSR training dataset with the same batch size. The performance are evaluated under the same training iterations.

\subsection{Comparisons with State-of-the-arts}
We evaluate UCIP with six state-of-the-art models on our CSR benchmarks which composes of five commonly adopted datasets: Set5~\cite{Set5}, Set14~\cite{Set14}, BSD100~\cite{BSDall}, Urban100~\cite{urban100} and Manga109~\cite{manga109}. The compared models include the fully-convolutional network RRDB~\cite{wang2018esrganRRDB}, the transformer-based image restoration model SwinIR~\cite{liang2021swinir} and its upgraded version Swin2SR~\cite{conde2022swin2sr}, the MLP-like model MAXIM~\cite{tu2022maxim} and two multi-task models AIRNet~\cite{Airnet} and PromptIR~\cite{promptir}. We add the HR reconstruction module to last three models, enabling them to perform super-resolution tasks. \textit{All compared methods
are trained from scratch with our proposed UCSR dataset for fair comparisons.}

\begin{table*}[htp]
\centering
\caption{Quantitative comparison for compressed image super-resolution on learning-based codecs. Results are tested on $\times4$ with different compression qualities in terms of  PSNR$\uparrow$/SSIM$\uparrow$. The best performances are in \tcr{red}. Notice that, as HIFIC~\cite{mentzer2020hifi} does not support some low-resolution images from downsampled Set5 and Set14 datasets, we do not use HIFIC codec to compress these two datasets. \textit{Notably, all compared methods are trained from scratch with our proposed UCSR dataset for fair comparisons.} $\mathcal{D}$ denotes for ``Datasets''.}
\setlength{\tabcolsep}{2pt}
\resizebox{\textwidth}{!}{
\begin{tabular}{c|c|@{\hspace{5pt}}c@{\hspace{5pt}}|cccc|cccc|ccc}
\hline
\multirow{2}{*}{$\mathcal{D}$} & \multirow{2}{*}{Methods} & \multirow{2}{*}{Params}  & \multicolumn{4}{c|}{$\text{C}_\text{PSNR}$~\cite{cheng2020learned}}                             & \multicolumn{4}{c}{$\text{C}_\text{SSIM}$~\cite{cheng2020learned}}  & \multicolumn{3}{c}{HIFIC~\cite{mentzer2020hifi}}                              \\ \cline{4-14} 
                          &                          &                               & $\mathcal{Q}=1 $          & $\mathcal{Q}=2 $          & $\mathcal{Q}=3 $          & $\mathcal{Q}=4 $          & $\mathcal{Q}=1 $          & $\mathcal{Q}=2 $          & $\mathcal{Q}=3 $          & $\mathcal{Q}=4 $    &     $\mathcal{Q}=\text{`low'}$  &  $\mathcal{Q}=\text{`med'}$  
    & $\mathcal{Q}=\text{`high'}$ \\ \hline
\multirow{7}{*}{\rotatebox{90}{Set5~\cite{Set5}}}     & RRDB~\cite{wang2018esrganRRDB}                     & 16.70M                       & 24.54/0.698 & 25.40/0.725 & 26.14/0.746 & 27.37/0.781 & 21.19/0.591 & 21.91/0.621 & 22.94/0.654 & 23.69/0.684 & - & - & -   \\
                          & SwinIR~\cite{liang2021swinir}                   & 11.72M                     & 24.55/0.704                    &     25.50/0.731        &     26.25/0.753        &       27.70/0.790      &         21.19/0.595    &      21.93/0.629       &      22.96/\tcr{0.663}       &    23.70/0.689         &   - & - &-          \\
                          & Swin2SR~\cite{conde2022swin2sr}                  & 12.05M                     & 24.56/0.702                    &      25.51/0.732       &       26.27/0.756      &        27.69/0.792     &       21.23/0.595      &       21.95/0.629      &      22.98/0.662       &    23.77/0.691         &  -&-&-           \\
                          & MAXIM~\cite{tu2022maxim}       &      26.74M       & 24.58/0.704                     &        25.49/0.733             &       26.27/0.755      &      27.66/0.791       &   21.26/0.595          &      \tcr{22.01}/0.631       &     \tcr{23.03/0.663}        &       \tcr{23.78/0.692}      &   -          &    - &-          \\
                          & AIRNet~\cite{Airnet}                   & 7.76M                     & 24.54/0.702                    &          25.41/0.730   &       26.16/0.751      &     27.50/0.789        &      21.15/0.595       &      21.91/0.629       &        22.93/0.660     &    23.67/0.689         &   -& - &-          \\
                          & PromptIR~\cite{promptir}                 & 35.72M                     & 24.48/0.700                    &         25.42/0.730    &        26.16/0.751     &      27.72/0.793       &     21.26/0.600        &       21.99/0.632      &       22.98/0.661      &     23.66/0.686        &  -&-&-           \\
                          & UCIP                & 11.42M                     & \tcr{24.65/0.705}                    &       \tcr{25.59/0.736}      &       \tcr{26.39/0.758}      &     \tcr{27.93/0.796}        &      \tcr{21.30/0.607}       &        21.98/\tcr{0.633}     &      23.01/\tcr{0.663}       &      23.74/0.689       &  - & - & -           \\ \hline\hline
\multirow{7}{*}{\rotatebox{90}{Set14~\cite{Set14}}}    & RRDB~\cite{wang2018esrganRRDB}                     &   16.70M                      &           23.52/0.588             &     24.17/0.611        &      24.76/0.632       &      25.51/0.662       &       21.19/0.526      &         21.70/0.545    &      22.38/0.564       &      23.04/0.586       &     - & - & -        \\
                          & SwinIR~\cite{liang2021swinir}                   &     11.72M                     & 23.61/0.591                        &      24.29/0.615       &     24.92/0.638        &       25.82/0.673      &        21.20/0.527     &      21.69/0.545       &     22.39/0.566        &       23.03/0.587      &   - & - & -          \\
                          & Swin2SR~\cite{conde2022swin2sr}                  &  12.05M                     & 23.61/0.591                     &       24.29/0.615      &       24.91/0.638      &       25.82/0.673      &       21.20/0.527      &       21.72/0.546      &       22.40/0.566      &           23.05/0.587  &  -&-&-           \\
                          & MAXIM~\cite{tu2022maxim}           &      26.74M          &   23.61/0.592                     &       24.31/0.615              &      24.94/0.639       &        25.82/0.673     &    21.23/0.527         &     \tcr{21.75}/0.546        &        22.42/0.565     &     \tcr{23.10}/0.588        &    -         &       -&-      \\
                          & AIRNet~\cite{Airnet}     
              &        7.76M                     & 23.53/0.590                           &        24.19/0.614     &       24.82/0.637      &      25.65/0.672       &     21.14/0.526        &     21.66/0.545        &       22.35/0.565      &        23.00/0.587     &   - &- &-          \\
                          & PromptIR~\cite{promptir}                 &  35.72M                     & 23.62/0.592                   &       24.30/0.616      &       24.95/0.639      &     25.85/0.675        &      21.20/0.528       &       21.73/\tcr{0.547}      &       \tcr{22.43}/0.566      &     23.08/0.588        &   - &- &-          \\
                          & UCIP                &  11.42M                     & \tcr{23.66/0.593}                       &      \tcr{24.34/0.617}       &       \tcr{25.00/0.641}      &     \tcr{25.97/0.678}        &       \tcr{21.24/0.529}      &        21.73/\tcr{0.547}     &        22.41/\tcr{0.567}     &       \tcr{23.10/0.590}      &    - & - & -         \\ \hline\hline
\multirow{7}{*}{\rotatebox{90}{BSD100~\cite{BSDall}}}    & RRDB~\cite{wang2018esrganRRDB}                     &    16.70M                     &           23.55/0.548             &      24.15/0.569       &      24.67/0.590       &       25.27/0.618      &       21.95/0.507      &       22.44/0.523      &       23.07/0.540      &       23.53/0.556      &    21.27/0.521   &      21.99/0.550   &     22.38/0.575   \\
                          & SwinIR~\cite{liang2021swinir}                   &    11.72M                     & 23.58/0.549                     &       24.19/0.573      &      24.73/0.595       &      25.42/0.627       &         21.95/0.508    &     22.45/0.524        &       23.07/0.541      &     23.54/0.557        & 21.46/0.531  &22.11/0.556 &   22.59/0.581       \\
                          & Swin2SR~\cite{conde2022swin2sr}                  &   12.05M                     & 23.57/0.550                 &     24.19/0.573        &    24.71/0.595         &       25.39/0.627      &      21.95/0.507       &         22.44/0.524    &      23.07/0.541       &       23.53/0.557      &      21.44/0.531    &22.12/0.557  & 22.55/0.582 \\
                          & MAXIM~\cite{tu2022maxim}          &      26.74M           &   23.59/0.550                     & 24.20/0.573                &       24.74/0.595      &       25.42/0.628      &        21.97/0.508     &      22.47/0.524       &        23.10/0.541     &      23.56/0.557       &        21.51/0.531     & 22.17/0.557 &      22.54/0.580      \\
                          & AIRNet~\cite{Airnet}                   &                      7.76M                     & 23.56/0.549                &       24.16/0.572      &      24.70/0.595       &       25.36/0.627      &       21.95/0.507      &     22.44/0.524        &        23.05/0.541     &       23.51/0.557      & 21.44/0.530  &22.11/0.556&     22.51/0.579     \\
                          & PromptIR~\cite{promptir}                 &   35.72M                     & 23.59/0.550                  &       24.21/0.573      &      24.75/0.596       &    25.43/0.628         &      21.96/0.508       &      22.47/0.524       &        23.09/0.541     &      23.55/0.558       &     21.49/0.532 &22.14/0.558 &22.66/0.583        \\
                          & UCIP                &   11.42M                     & \tcr{23.61/0.551}                       &       \tcr{24.24/0.575}      &      \tcr{24.77/0.597}       &       \tcr{25.49/0.630}        &    \tcr{21.98}/0.508       &     \tcr{22.48/0.525}       &         \tcr{23.12/0.543}    &       \tcr{23.59/0.560}      &    \tcr{21.94/0.534}    &  \tcr{22.19/0.559} &  \tcr{23.39/0.587}   \\ \hline\hline
\multirow{7}{*}{\rotatebox{90}{Urban100~\cite{urban100}}}    & RRDB~\cite{wang2018esrganRRDB}                     &      16.70M                   &            21.70/0.580           &      22.21/0.603       &      22.60/0.622       &      23.21/0.654       &    19.84/0.504         &       20.29/0.524      &      20.81/0.546       &          21.31/0.570   &  20.70/0.540     & 21.42/0.570&   21.89/0.593    \\
                          & SwinIR~\cite{liang2021swinir}                   &  11.72M                     & 21.78/0.587                    &      22.32/0.613       &      22.76/0.633       &        23.52/0.672     &   19.87/0.509          &     20.31/0.530        &      20.86/0.553       &    21.36/0.579         &   20.83/0.549 & 21.60/0.579&    22.11/0.605     \\
                          & Swin2SR~\cite{conde2022swin2sr}                  &   12.05M                     &       21.82/0.588          &     22.38/0.614        &       22.80/0.635      &        23.52/0.672     &      19.89/0.510       &          20.33/0.531   &      20.89/0.554       &      21.41/0.580       &     20.87/0.550  & 21.65/0.580  & 22.16/0.607    \\
                          & MAXIM~\cite{tu2022maxim}        &      26.74M             &  21.80/0.587                     & 22.34/0.612                &     22.77/0.633        &      23.49/0.670       &      19.90/0.509       &     20.35/0.530        &           20.89/0.553  &      21.40/0.578       &     20.87/0.549        &  21.63/0.579   &   22.14/0.605     \\
                          & AIRNet~\cite{Airnet}                   &                    7.76M                     & 21.71/0.585                 &      22.23/0.610       &     22.65/0.631        &      23.34/0.668       &         19.84/0.507    &     20.27/0.527        &       20.81/0.550      &          21.30/0.575   &  20.77/0.546 & 21.49/0.576  &    21.98/0.600      \\
                          & PromptIR~\cite{promptir}                 &  35.72M                     & 21.82/0.589                   &        22.35/0.614     &      22.79/0.635       &      23.53/0.673       &    19.92/0.511         &       20.36/0.532      &       20.91/0.555      &        21.42/0.581     & 20.88/0.552 & 21.64/0.583&      22.19/0.610     \\
                          & UCIP                &  11.42M                     & \tcr{21.89/0.593}                       &     \tcr{22.46/0.620}        &       \tcr{22.91/0.641}      &     \tcr{23.72/0.682}        &       \tcr{19.98/0.516}      &      \tcr{20.43/0.538}       &        \tcr{21.01/0.563}     &      \tcr{21.56/0.590}       & \tcr{20.95/0.555}   & \tcr{21.77/0.588}& \tcr{22.30/0.616}         \\ \hline\hline
\multirow{7}{*}{\rotatebox{90}{Manga109~\cite{manga109}}}    & RRDB~\cite{wang2018esrganRRDB}                     &     16.70M                    &       23.20/0.725                 &       23.90/0.746      &      24.43/0.762       &       25.47/0.794      &        20.12/0.635     &      20.75/0.657       &     21.55/0.680        &     22.27/0.705        &   21.28/0.671  & 22.41/0.705 &  23.10/0.729       \\
                          & SwinIR~\cite{liang2021swinir}                   &     11.72M                     & 23.31/0.733                &         24.13/0.760    &     24.68/0.775        &        26.01/0.813     &      20.12/0.641       &     20.75/0.664        &        21.58/0.689     &          22.34/0.715   &     21.32/0.680 &22.55/0.716 &   23.31/0.742      \\
                          & Swin2SR~\cite{conde2022swin2sr}                  &  12.05M                     & 23.33/0.733               &      24.08/0.757       &      24.68/0.774       &        25.91/0.809     &       20.15/0.641      &           20.79/0.664  &      21.62/0.689       &       22.38/0.715      &        21.36/0.679  & 22.61/0.715 & 23.37/0.742   \\
                          & MAXIM~\cite{tu2022maxim}       &      26.74M              &   23.31/0.732                     & 24.08/0.756                 &       24.67/0.774      &     25.95/0.810        &    20.15/0.640         &      20.78/0.663       &        21.60/0.687     &       22.34/0.712      &   21.36/0.679          &   22.58/0.715  &      23.33/0.741   \\
                          & AIRNet~\cite{Airnet}                   &                        7.76M                     & 23.17/0.729             &       23.89/0.752      &      24.47/0.770       &      25.66/0.807       &        20.08/0.637     &       20.70/0.660      &       21.49/0.684      &          22.22/0.709   & 21.33/0.676  &22.42/0.711 &  23.18/0.736        \\
                          & PromptIR~\cite{promptir}                 &   35.72M                     & 23.33/0.734                    &      24.09/0.758       &     24.70/0.775        &    25.97/0.812         &        20.15/0.642     &       20.79/0.665      &       21.62/0.690      &     22.36/0.715        &    21.39/0.682&22.58/0.718 &  23.40/0.745       \\
                          & UCIP                &  11.42M                     & \tcr{23.43/0.737}                     &       \tcr{24.24/0.762}      &      \tcr{24.88/0.781}       &      \tcr{26.29/0.819}       &      \tcr{20.19/0.645}       &       \tcr{20.85/0.669}      &         \tcr{21.73/0.696}    &      \tcr{22.53/0.723}      &   \tcr{21.41/0.684}    & \tcr{22.70/0.723} &    \tcr{23.55/0.750}  \\ \hline
\end{tabular}
}
\label{table:learning-based}
\end{table*}

As demonstrated in Table.~\ref{table:traditional} and Table.~\ref{table:learning-based}, our UCIP outperforms all other methods on almost all codecs and compression qualities. Particularly, UCIP achieves PSNR gain of up to 0.45dB against PromptIR~\cite{promptir} with only one-third the number of parameters. Another intriguing observation is that the gains provided by UCIP become more significant as the compression ratio decreases. We attribute this to the preservation of more high-frequency information at milder compression levels. The abundance of high-frequency details further enhances the capability of PTMM to conduct global-wise informative tokens extraction, thus leads to a better performance.

\begin{figure*}[htp]
    \centering
    \includegraphics[width=0.98\linewidth]{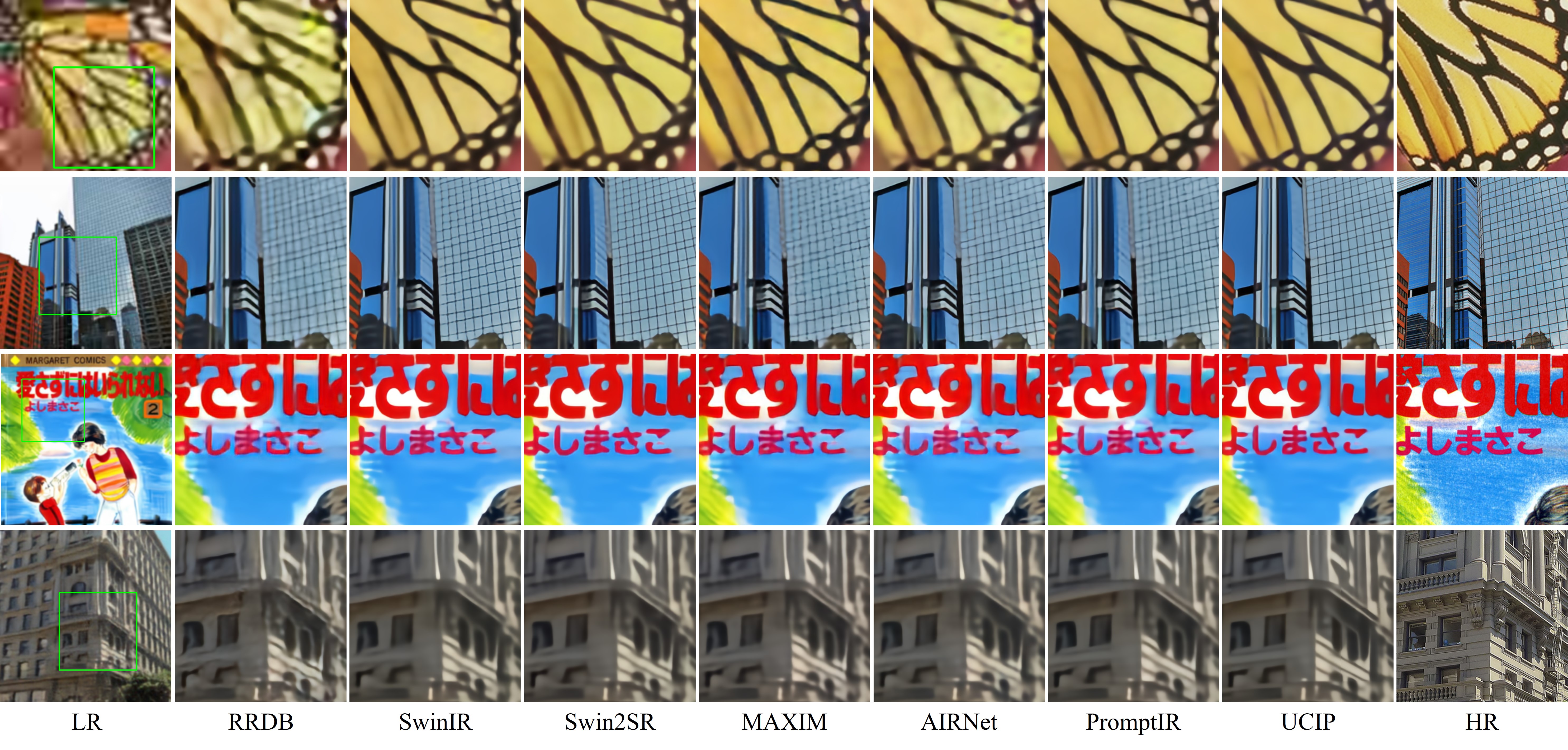}
    \caption{Visual comparisons between UCIP and other state-of-the-art methods. To demonstrate the effectiveness of UCIP across different codecs, we display four rows of images, each compressed with JPEG($\mathcal{Q}=10$), HM($\mathcal{Q}=32$), $\text{C}_{\text{PSNR}}$($\mathcal{Q}=4$) and HIFIC($\mathcal{Q}=\text{`med'}$), respectively. We show more results in the Sec.~\ref{more_sub}.}
    \label{fig:all}
\end{figure*}

As illustrated in Fig.~\ref{fig:all}, UCIP leverages the implicit guidance of the dynamic prompt to recover more textural details while avoiding the generation of artifacts. Specifically, as observed in the first row, our model recovers the clearest texture of the monarch. Besides, in the second and final rows, images reconstructed by our method exhibit clearer edges and fewer distorted lines. For the third row, our method successfully removes compression artifacts, while other methods suffer from blocked and blurry outputs. We attribute these performances to the generation of the dynamic prompt and the fusion of global tokens with local features.

\begin{figure*}[htp]
    \centering
    \includegraphics[width=1.0\linewidth]{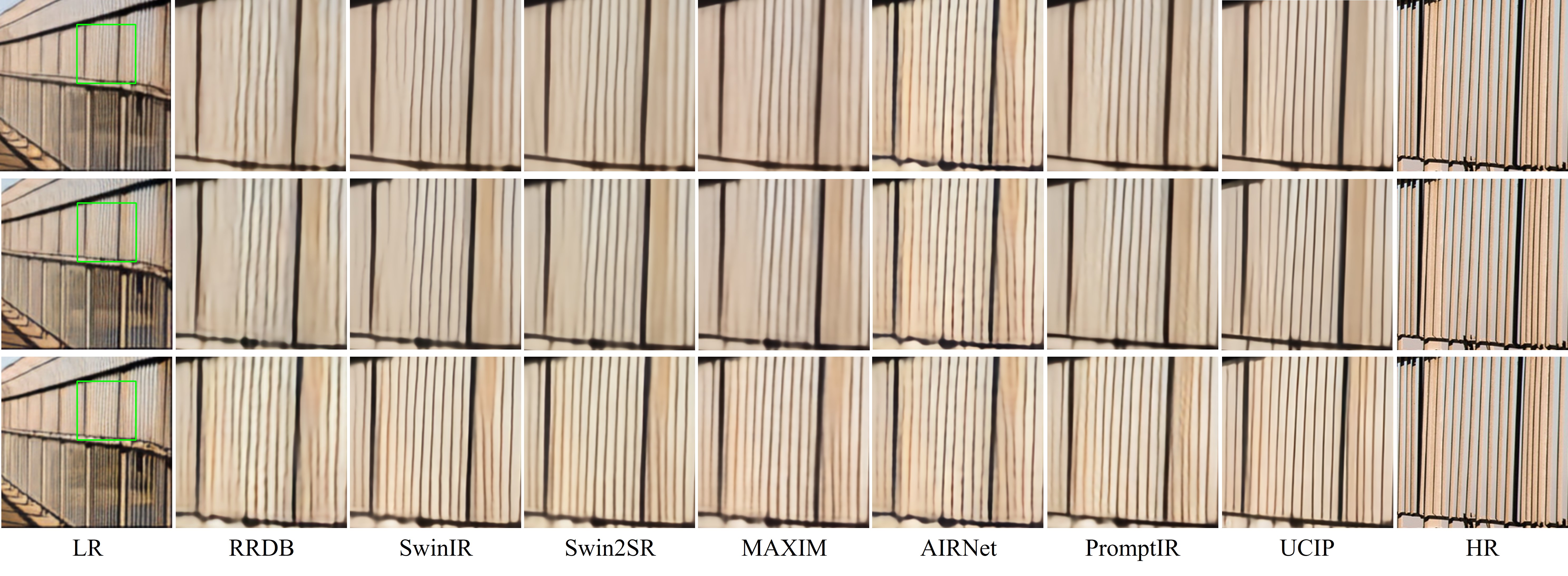}
    \caption{Visual comparisons between UCIP and other state-of-the-art methods under different compression qualities within HIFIC~\cite{mentzer2020hifi} codec. The qualities of HIFIC from top to bottom are `low', `medium', and `high', respectively. We show more results in the Sec.~\ref{more_sub}.}
    \label{fig:hifiall}
\end{figure*}

It is noteworthy that, though we do not specifically tailor prompts for various compression qualities within certain codec, experimental evidence suggests that our dynamic prompt not only possesses task-specific adaptability but is also capable of handling different distortion degrades. As shown in Fig.~\ref{fig:hifiall}, our method maintains robust image restoration capabilities across three levels of compression qualities (\eg, always recovers straight lines on the right side of image) .

\begin{table*}[htp]
\centering
\caption{Quantitative comparison for different tuning methods on two new codecs. Results are tested on $\times4$ with different compression qualities in terms of  PSNR$\uparrow$/SSIM$\uparrow$. \textbf{The first line of each benchmark denotes the baseline model trained with our proposed UCSR dataset, which is directly evaluated on these new codecs.} Notice that, for both codecs, the smaller the quality factor, the poorer the image quality. The best performances are in \tcr{red}. Zoom in for best view.}
\setlength{\tabcolsep}{2pt}
\resizebox{\textwidth}{!}{
\begin{tabular}{c|c|@{\hspace{5pt}}c@{\hspace{5pt}}|c|cccc|cccc}
\hline
\multirow{2}{*}{Datasets} & Pre-train & Add prompts & Fine-tune & \multicolumn{4}{c|}{WebP~\cite{webp}}                             & \multicolumn{4}{c}{ELIC~\cite{he2022elic}}                               \\  \cline{5-12} 
                          &                          with prompts & in fine-tune & which part                                & $\mathcal{Q}=10 $          & $\mathcal{Q}=20 $          & $\mathcal{Q}=30 $          & $\mathcal{Q}=40 $          & $\mathcal{Q}=1 $          & $\mathcal{Q}=2 $          & $\mathcal{Q}=3 $          & $\mathcal{Q}=4 $     \\ \hline
\multirow{4}{*}{Set5}    & -                     & -               &    -   & 26.20/0.750 & 27.12/0.777 & 27.78/0.794 & 28.33/0.806 & \tcr{22.48}/0.643 & 23.51/0.677 & 24.63/0.708 & 25.55/0.735   \\ 
& \ding{55}                   & \ding{51}               &    only prompts   & 26.25/0.753 & 27.16/0.778 & 27.83/0.795 & 28.40/0.809 & 22.43/0.639 & \tcr{23.57}/0.678 & 24.81/0.713 & 25.97/0.745   \\
                          & \ding{55}                & \ding{51}                    & full model &\tcr{26.55/0.763}                    &     \tcr{27.49/0.787}        &     \tcr{28.20/0.805}        &       \tcr{28.79/0.818}      &         22.41/0.642    &      23.53/0.678       &      24.83/\tcr{0.716}       &    \tcr{26.06/0.749}                \\
                          & \ding{51}                  & \ding{51}  &    only prompts              & 26.54/0.762 & \tcr{27.49/0.787} & 28.18/0.804 & 28.78/\tcr{0.818}     &       22.42/\tcr{0.644} & 23.55/\tcr{0.679} & \tcr{24.84}/0.715 & 26.03/0.748                 \\
\hline\hline
\multirow{4}{*}{Set14}  & -                     & -               &    -   & 24.69/0.631 & 25.41/0.658 & 25.87/0.675 & 26.25/0.689 & 21.84/0.539 & 22.78/0.566 & 23.62/0.593 & 24.37/0.618   \\ 
& \ding{55}                   & \ding{51}               &    only prompts   & 24.76/0.634 & 25.43/0.659 & 25.91/0.678 & 26.29/0.693 & \tcr{21.86}/0.540 & 22.81/0.567 & 23.69/0.595 & 24.58/0.625   \\
                          & \ding{55}                & \ding{51}                    & full model & 24.93/0.641                    &     25.64/0.667        &     26.12/\tcr{0.686}       &       26.53/\tcr{0.701}     &         21.85/\tcr{0.541}    &      22.81/\tcr{0.569}       &      \tcr{23.73/0.597}       &    24.67/0.628                 \\
                          & \ding{51}                  & \ding{51}  &    only prompts              & \tcr{24.97/0.642} & \tcr{25.65/0.668} & \tcr{26.14/0.686} & \tcr{26.55/0.701}     &       21.84/0.540 & 22.81/0.568 & \tcr{23.73/0.597} & \tcr{24.69/0.629}                \\
                          \hline\hline
\multirow{4}{*}{BSD100}  & -                     & -               &    -   & 24.36/0.583 & 24.93/0.608 & 25.34/0.626 & 25.63/0.639 & \tcr{22.14}/0.507 & \tcr{22.90}/0.528 & 23.61/0.551 & 24.20/0.574   \\ 
& \ding{55}                   & \ding{51}               &    only prompts   & 24.44/0.587 & 25.01/0.612 & 25.42/0.630 & 25.74/0.644 & 22.12/0.508 & 22.89/0.529 & 23.63/0.553 & 24.33/0.580   \\
                          & \ding{55}                & \ding{51}                    & full model &\tcr{24.55}/0.592                    &     \tcr{25.11}/0.616        &     25.52/0.635        &       25.85/0.649      &         22.10/\tcr{0.509}    &      22.88/\tcr{0.530}      &      \tcr{23.64/0.554}      &    24.36/0.581                 \\
                          & \ding{51}                  & \ding{51}  &    only prompts              & \tcr{24.55/0.593} & \tcr{25.11/0.617} & \tcr{25.53/0.636} & \tcr{25.86/0.650}     &       22.10/\tcr{0.509} & 22.89/\tcr{0.530} & \tcr{23.64/0.554} & \tcr{24.37/0.582}                 \\
                          \hline\hline
\multirow{4}{*}{Urban100}  & -                     & -               &    -   & 22.81/0.640 & 23.43/0.669 & 23.81/0.687 & 24.07/0.698 & 20.49/0.533 & 21.32/0.569 & 22.00/0.602 & 22.54/0.629   \\ 
& \ding{55}                   & \ding{51}               &    only prompts   & 22.72/0.634 & 23.30/0.661 & 23.67/0.679 & 23.94/0.691 & 20.48/0.532 & 21.31/0.568 & 22.05/0.601 & 22.70/0.632   \\
                          & \ding{55}                & \ding{51}                    & full model &23.06/0.653                    &     23.66/0.680        &     24.05/0.698        &       24.34/0.710      &         20.47/0.535    &      21.33/0.572       &      22.10/0.607       &    22.81/0.640                 \\
                          & \ding{51}                  & \ding{51}  &    only prompts              & \tcr{23.15/0.656} & \tcr{23.74/0.683} & \tcr{24.13/0.701} & \tcr{24.41/0.713}     &       \tcr{20.54/0.538} & \tcr{21.40/0.575} & \tcr{22.18/0.610} & \tcr{22.88/0.643}                \\
                          \hline\hline
\multirow{4}{*}{Manga109}  & -                     & -               &    -   & 24.77/0.779 & 25.69/0.805 & 26.28/0.820 & 26.70/0.831 & 21.20/0.671 & 22.36/0.707 & 23.36/0.739 & 24.22/0.766   \\ 
   & \ding{55}                   & \ding{51}               &    only prompts   & 24.76/0.776 & 25.65/0.801 & 26.23/0.817 & 26.66/0.828 & 21.22/0.671 & 22.42/0.708 & 23.52/0.741 & 24.60/0.774   \\
                          & \ding{55}                & \ding{51}                    & full model &25.09/0.789                    &     26.01/0.813        &     26.65/0.829       &       27.16/0.841     &         21.19/0.670    &      22.42/0.711       &      23.58/0.746       &    24.73/0.780                \\
                          & \ding{51}                  & \ding{51}  &    only prompts             & \tcr{25.18/0.791} & \tcr{26.11/0.815} & \tcr{26.75/0.831} & \tcr{27.25/0.843}     &       \tcr{21.23/0.675} & \tcr{22.48/0.713} & \tcr{23.65/0.747} & \tcr{24.82/0.783}                 \\

\hline
\end{tabular}
}
\label{table:finetune}
\end{table*}

\subsection{Prompt Tuning for UCIP}
Prompt learning can be utilized in two popular ways: (i) One is to utilize prompt learning for multi-task learning, \eg, PromptIR~\cite{promptir}, ProRes~\cite{ma2023prores}, PIP~\cite{PIP} in low-level vision, which needs to train whole model from scratch; (ii) another is prompt tuning, which requires a strong baseline model and aims to optimize only a small part of parameters for downstream tasks. Notably, in the CSR field, there are no pre-trained baseline models on multiple types of compression artifacts existed, which prevents us to study prompt tuning in the beginning. And thus, we build the first Universal CSR framework and corresponding dataset with the first way, which follows existing prompt learning works in low-level vision~\cite{promptir,ma2023prores,PIP,luo2023controlling}. However, training a model from scratch is time consuming. To further explore the potential of our proposed UCIP in prompt tuning, we choose two \textit{unseen} codecs, including one traditional codec WebP~\cite{webp} and one learning-based codec ELIC~\cite{he2022elic}, to fine-tune UCIP. In Tab.~\ref{table:finetune}, we explore four ways of fine-tuning: i) directly evaluated pre-trained model without fine-tuning. ii) pre-training without prompt, then adding prompt and only training prompt parameters on new tasks. iii) pre-training without prompt, then adding prompt and only training prompt parameters on new tasks. iv) pre-training with prompt, then fine-tuning only prompt parameters on new tasks. All the experiments are conducted under the same settings with the same training iterations. As shown in Tab.~\ref{table:finetune}, compared between ii) and iii), tuning only prompt achieves comparable performance on ELIC codec against tuning full model. Compared between iii) and iv), tuning only prompt based on UCIP achieves comparable and even better performances against tuning full model after adding prompt. The experimental results indicate that our proposed UCIP can serve as a strong baseline model in CSR field, which will also benefit the prompt tuning for new codecs in future works.

\subsection{Ablation Studies}
\subsubsection{The effects of dynamic prompt}
To validate the effectiveness of our DPM, we conduct experiments on the different prompt designs. The results are shown in Table.~\ref{table:ab_prompt}. Specifically, without the dynamic prompt, UCIP is unable to perform task-wise informative token selection. Moreover, the use of fixed prompts may even impair the performance of UCIP, as they could provide incorrect guidance during the token mixing process. Compared to PromptIR~\cite{promptir}, our DPM utilizes very few parameters to achieve the spatial-adaptive modulation for tasks by only a few basic dynamic prompt kernels, thereby achieving a PSNR gain of up to 0.22dB. 

\begin{table}[htp]
\centering
\caption{Impacts of different prompt generation strategies. Results are reported on Manga109~\cite{manga109}. Flops is calculated based on input with the size $1\times3\times64\times64$.}
\resizebox{0.8\linewidth}{!}{\begin{tabular}{c|c|c|ccc}
\hline
\multirow{2}{*}{Method} & \multirow{2}{*}{Params(M)} & \multirow{2}{*}{Flops(G)}& \multicolumn{3}{c}{Codecs} \\ \cline{4-6} 
 & & & JPEG($\mathcal{Q}=40$) & HM($\mathcal{Q}=32$) & HIFIC($\mathcal{Q}=\text{`high'}$) \\ \hline
w/o Prompt & - & - & 25.70/0.805 & 27.22/0.840 & 23.39/0.745 \\
Fixed & - & - & 25.62/0.799 & 27.10/0.839 & 23.05/0.723 \\
PromptIR~\cite{promptir} & 9.66 & 0.900 & 25.89/0.810 & 27.42/0.845 &  23.50/0.749\\
Ours  & 0.46 & 0.024 & 26.11/0.813 & 27.61/0.848 & 23.55/0.750 \\ \hline
\end{tabular}}
\label{table:ab_prompt}
\end{table}

\subsubsection{The effects of local feature extraction}

As demonstrated in Sec.~\ref{sec:dpmb}, local feature extraction is essential for the model to aggregate useful local information with the content/spatial-aware task-adaptive contextual information. To validate this point, we conduct an ablation which replaces the local convolution with the identity module. As shown in Table.~\ref{table:ab_local}, PSNR drops about 0.1dB without local feature extraction, which indicates that incorporate global tokens with local features are necessary for CSR tasks.

\begin{table}[htp]
\parbox[t]{.44\linewidth}{
\centering
\caption{Impacts of local feature extraction. Results are reported on Manga109~\cite{manga109}.}
\resizebox{\linewidth}{!}{
\setlength{\extrarowheight}{4.4pt}
\begin{tabular}{c|cc}
\hline
\multirow{2}{*}{Codecs} & \multicolumn{2}{c}{Methods} \\ \cline{2-3} 
 & w/o $\operatorname{Conv}_{3\times 3}$ & Ours \\ \hline
JPEG($\mathcal{Q}=40$) & 25.97/0.809 & 26.11/0.813 \\
HM($\mathcal{Q}=32$) & 27.46/0.844 & 27.61/0.848 \\
HIFIC($\mathcal{Q}=\text{`high'}$) & 23.48/0.747 &  23.55/0.750 \\ \hline
\end{tabular}}
\label{table:ab_local}}
\hfill
\parbox[t]{.54\linewidth}{
\centering
\caption{Experiments about number of the dynamic prompt. Results are reported on Manga109~\cite{manga109}.}
\resizebox{\linewidth}{!}{
\begin{tabular}{c|ccc}
\hline
\multirow{2}{*}{Number} & \multicolumn{3}{c}{Codecs} \\ \cline{2-4} 
 & JPEG($\mathcal{Q}=40$) & HM($\mathcal{Q}=32$) & HIFIC($\mathcal{Q}=\text{`high'}$) \\ \hline
1 & 25.84/0.807 & 27.42/0.844 & 23.46/0.747 \\
2 & 25.88/0.808 & 27.42/0.845 & 23.48/0.748 \\
4 & 25.99/0.811 & 27.53/0.847 & 23.52/0.748 \\
8(Ours) & 26.11/0.813 & 27.61/0.848 & 23.55/0.750 \\
16 & 26.19/0.815 & 27.66/0.850 & 23.59/0.751 \\ \hline
\end{tabular}}
\label{table:ab_diversity}}
\end{table}

\subsubsection{The effects of number of the dynamic prompt}

To mine the content/spatial-aware task-adaptive contextual information for the universal CSR task, we introduce the dynamic prompt. In this part, we investigate the optimal number of the dynamic prompt. As demonstrated in Table~\ref{table:ab_diversity}, there is a noticeable constraint on the dynamic capacity of prompts for spatial content interpretation and degradation handling when the number of the dynamic prompt is small. As the number incrementally increases, the observed performance gap narrows, falling below our expectations. We attribute this to the inadequate weighting from input image features, primarily due to the constrained capabilities of a singular MLP layer. To strike a balance between performance and computational efficiency, we choose 8 as the number of the dynamic prompt.

\section{Conclusion}

In this paper, we present UCIP, the first universal Compressed Image Super-resolution model, which leverages a novel dynamic prompt structure with multi-layer perception (MLP)-like framework. Distinct from existing CSR works focused on a single compression codec JPEG, UCIP effectively addresses hybrid distortions across a spectrum of codecs. By utilizing the prompt-guided token mixer block (PTMB), it dynamically identifies and refines the content/spatial-aware task-adaptive contextual information, optimizing for different tasks and distortions. Our extensive experiments on the proposed comprehensive UCSR benchmarks confirm that UCIP not only achieves state-of-the-art performance but also demonstrates remarkable versatility and applicability. In future work, we will exploit the potential of UCIP and further improve both objective and subjective performances on UCSR benchmarks.

\section*{Acknowledgement}
This work was supported in part by NSFC under Grant 623B2098, 62021001, and 62371434. This work was mainly completed before March 2024.

%
%
\bibliographystyle{splncs04}
\bibliography{main}

\section*{Appendix}

\noindent Section~\ref{more_offset} illustrates the distribution of offsets from different PTMMs and codecs.

\noindent Section~\ref{more_sub} presents more qualitative results on various compression codecs and qualities.

\section{Distribution of Offsets}
\label{more_offset}
We investigate the learned distributions of offsets via histogram visualization of offsets from different Prompt guided Token Mixer Modules (PTMMs). The $i$ and $j$ of $\operatorname{PTMM}\_{i}\_{j}$ denotes the offsets of $j^{th}$ PTMM from $i^{th}$ PTMB. We have the following observations: 1) As the depth increases, the learned offsets first expand to a larger range and then shrink to a smaller range. This hints that the model learns to extract local information for the query token at shallow layers. In the middle layers, the model leverages the offsets to aggregate the global-wise information to perform better token mixing. At the last layers, the distortions contained in image features are mostly removed, therefore the model focuses more on using local information again to refine the query tokens for the reconstruction purpose. 2) The distribution of offsets from middle layers differs among various codecs. We attribute this to the guidance from task-specific prompts. Since the distortion varies among different codecs, the visualization of learned offsets validates that our prompts are capable of providing adaptive guidance against various distortions, thus leading to better performance in the CSR tasks~\cite{yang2023aim2022,li2022hst}. 3) The offsets expand to a wider range for learning-based codecs compared to traditional codecs. We believe this is because the distortion introduced by learning-based codecs is more challenging to eliminate compared to that from traditional codecs, necessitating broader ranges of offsets to extract useful information for query tokens.

\begin{figure*}
    \centering
    \begin{subfigure}{\textwidth}
        \centering
        \includegraphics[width=\linewidth]{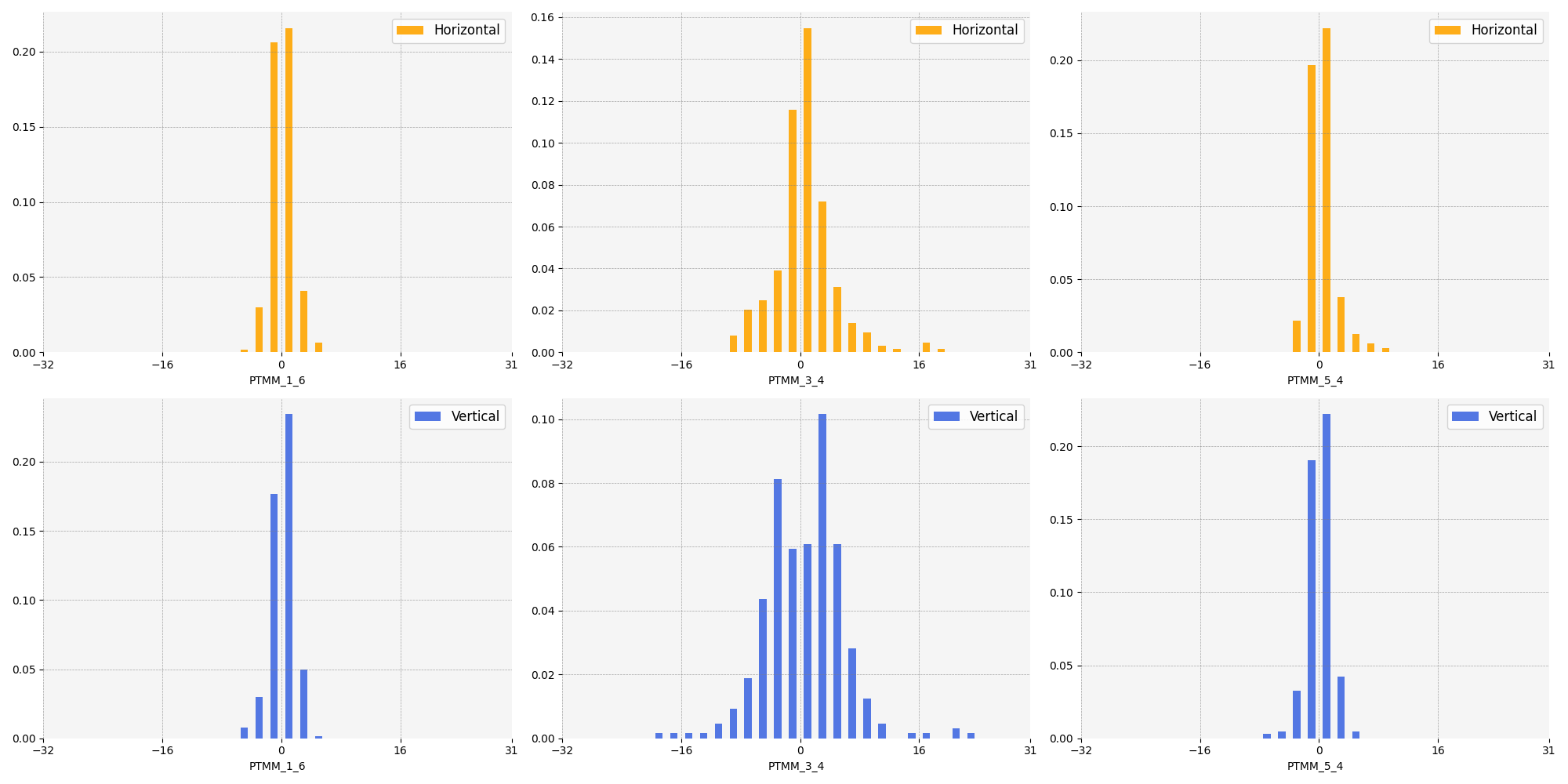}
        \caption{Visualization of learned offsets for JPEG~\cite{JPEG}.}
    \end{subfigure}

    \begin{subfigure}{\textwidth}
        \centering
        \includegraphics[width=\linewidth]{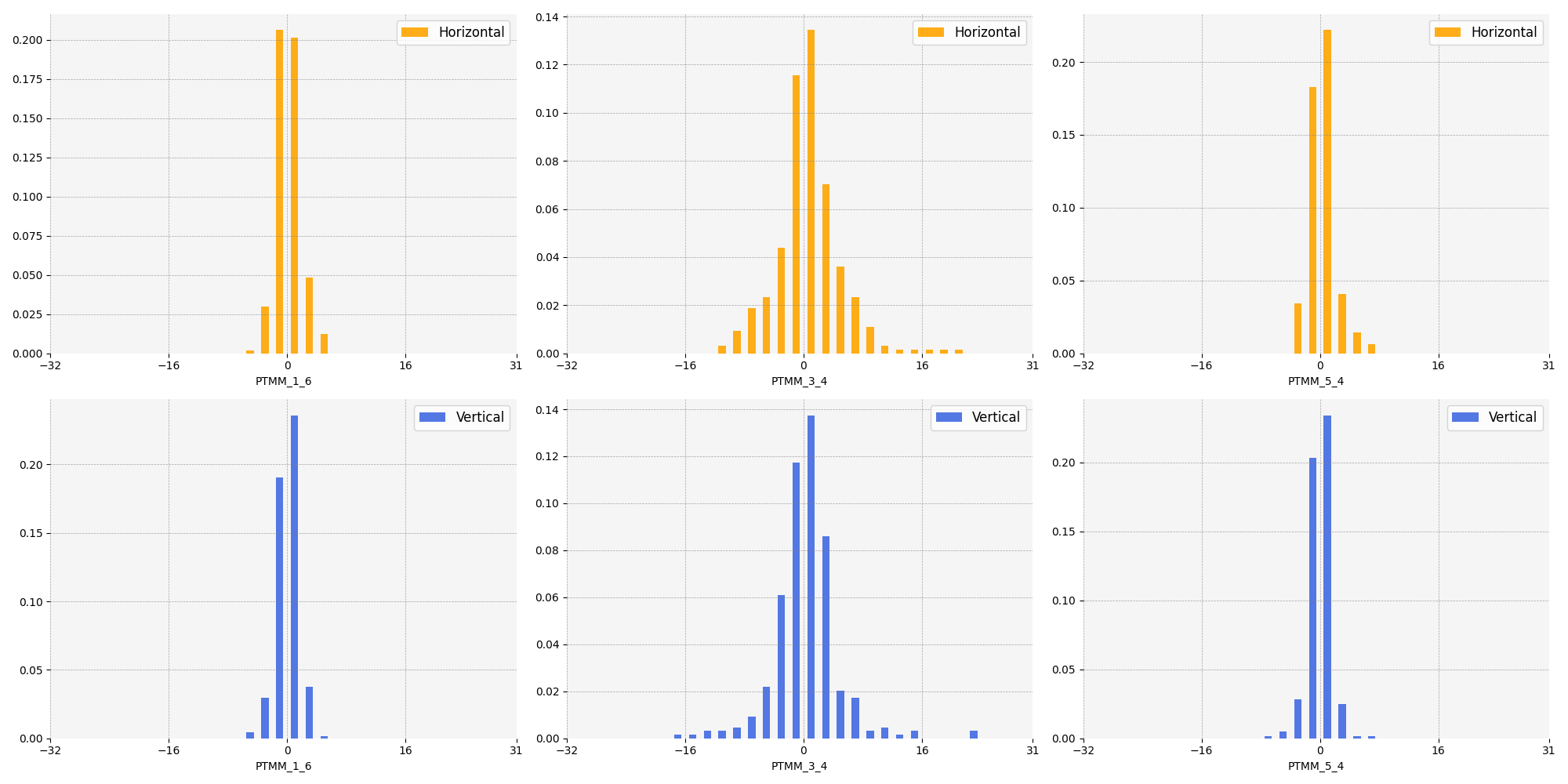}
        \caption{Visualization of learned offsets for VTM~\cite{VTM}.}
    \end{subfigure}
    \caption{Histograms of learned offsets for the center token from different PTMMs. LR images are randomly sampled from Urban100~\cite{urban100} compressed by $\text{C}_\text{PSNR}$~\cite{cheng2020learned} and VTM~\cite{VTM} codecs.}
\end{figure*}

\begin{figure*}
    \centering
    \begin{subfigure}{\textwidth}
        \centering
        \includegraphics[width=\linewidth]{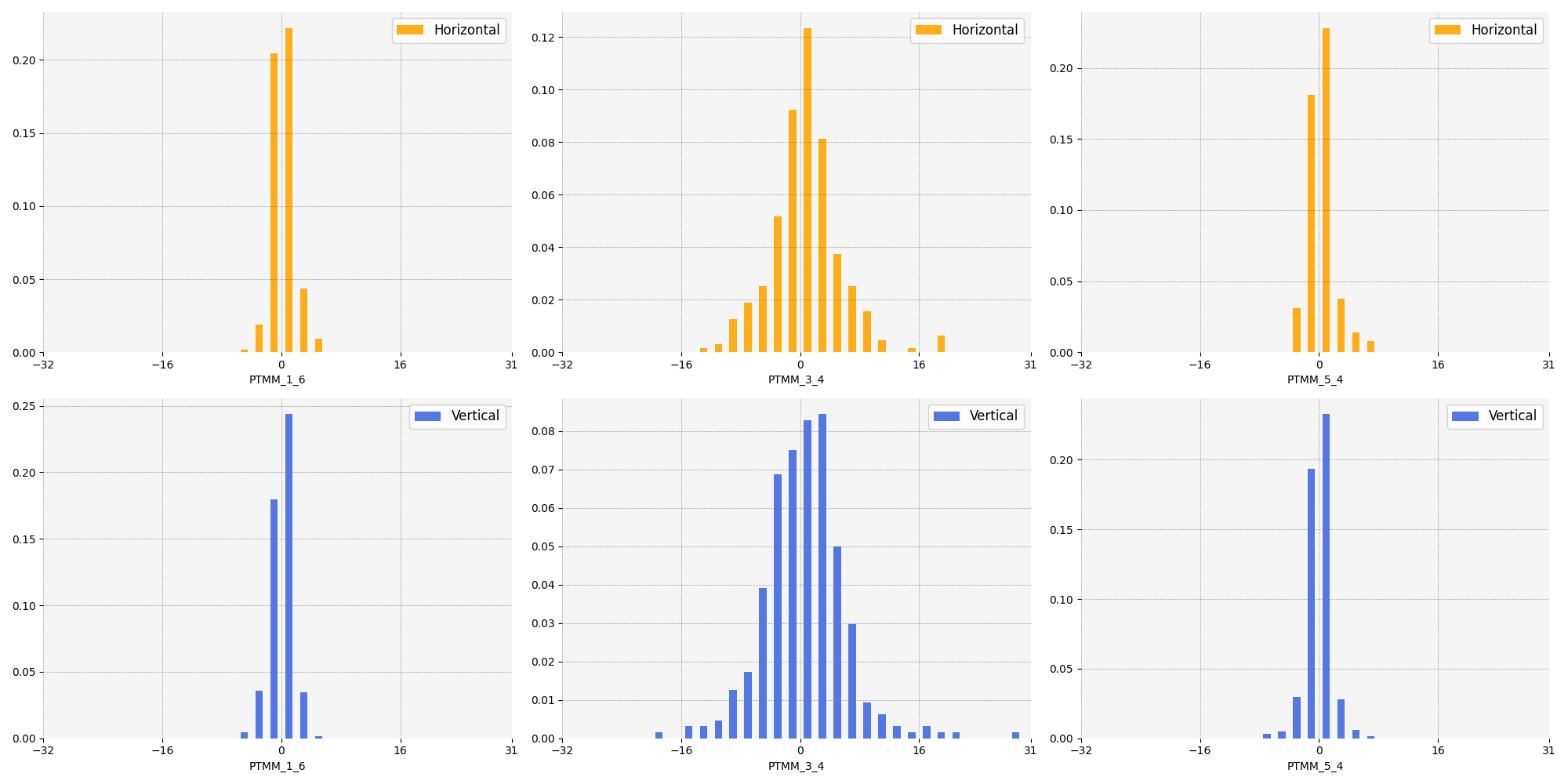}
        \caption{Visualization of learned offsets for $\text{C}_\text{PSNR}$~\cite{cheng2020learned}.}
    \end{subfigure}

    \begin{subfigure}{\textwidth}
        \centering
        \includegraphics[width=\linewidth]{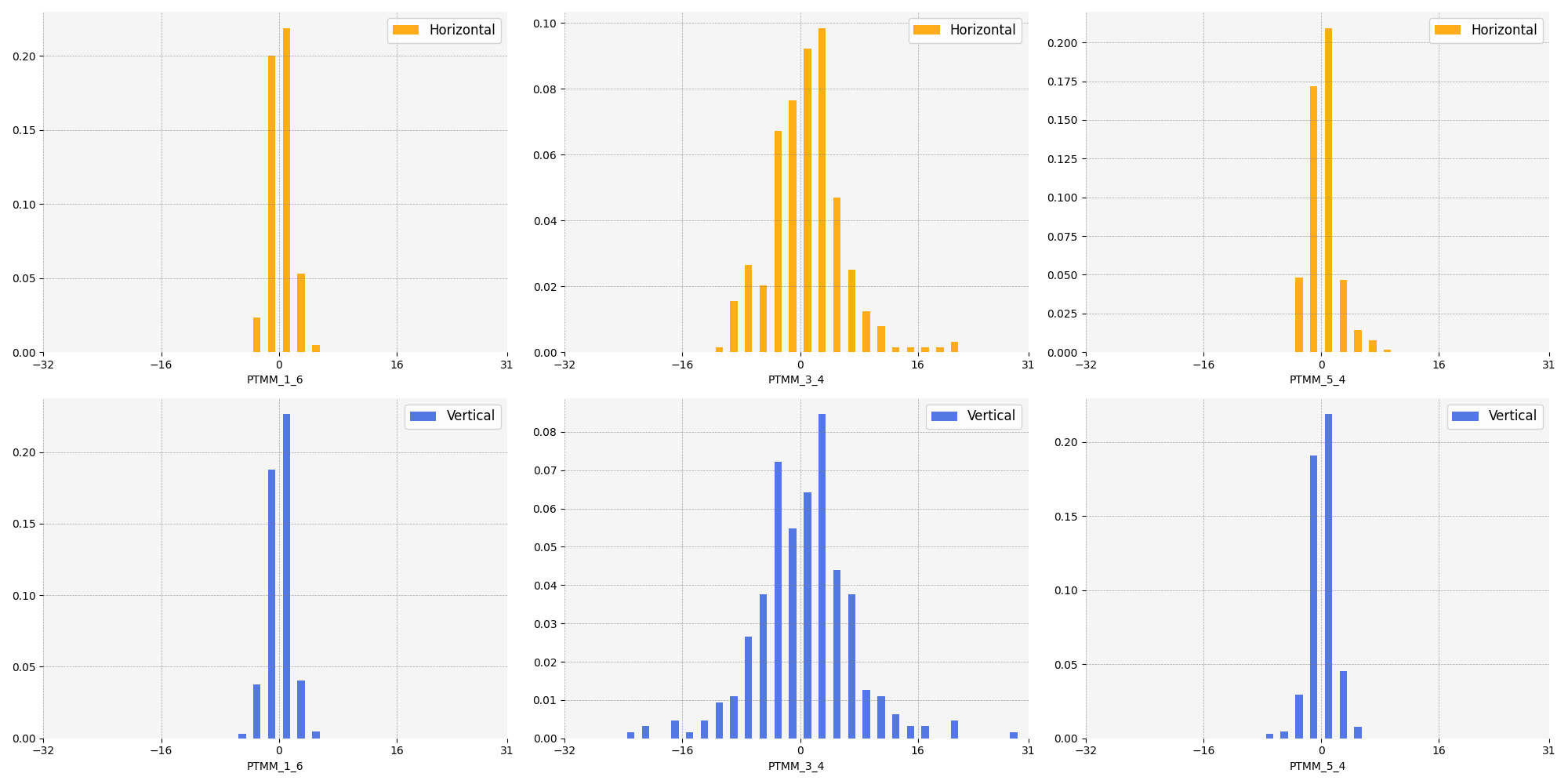}
        \caption{Visualization of learned offsets for HIFIC~\cite{mentzer2020hifi}.}
    \end{subfigure}
    \caption{Histograms of learned offsets for the center token from different PTMMs. LR images are randomly sampled from Urban100~\cite{urban100} compressed by JPEG~\cite{JPEG} and HIFIC~\cite{mentzer2020hifi} codecs.}
\end{figure*}

\section{More Visual Results}
\label{more_sub}
We provide more visual comparisons between our UCIP with state-of-the-art methods on different codecs and different compression qualities within a single codec. UCIP shows clearer textures and less artifacts in super-resolved images, indicating that our prompts and offsets are adaptive and robust against various degradations.

\begin{figure*}
    \centering
    \begin{subfigure}{.33\linewidth}
        \centering
        \includegraphics[width=0.95\linewidth]{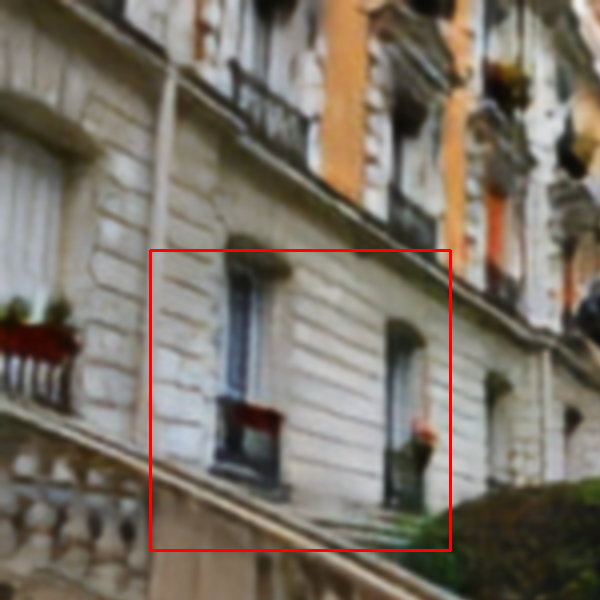}
        \caption*{Bicubic}
    \end{subfigure}%
    \begin{subfigure}{.33\linewidth}
        \centering
        \includegraphics[width=0.95\linewidth]{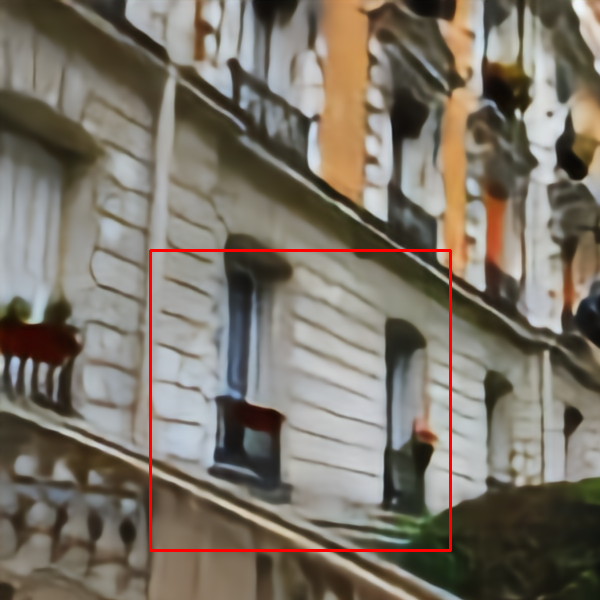}
        \caption*{RRDB~\cite{wang2018esrganRRDB}}
    \end{subfigure}%
    \begin{subfigure}{.33\linewidth}
        \centering
        \includegraphics[width=0.95\linewidth]{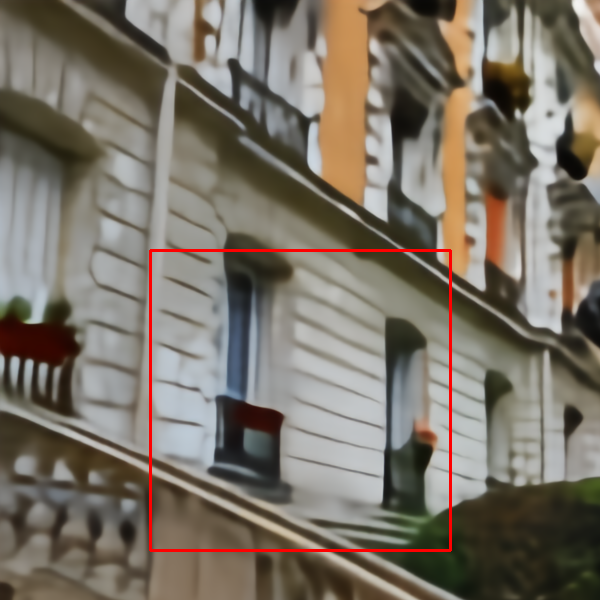}
        \caption*{SwinIR~\cite{liang2021swinir}}
    \end{subfigure}
    \begin{subfigure}{.33\linewidth}
        \centering
        \includegraphics[width=0.95\linewidth]{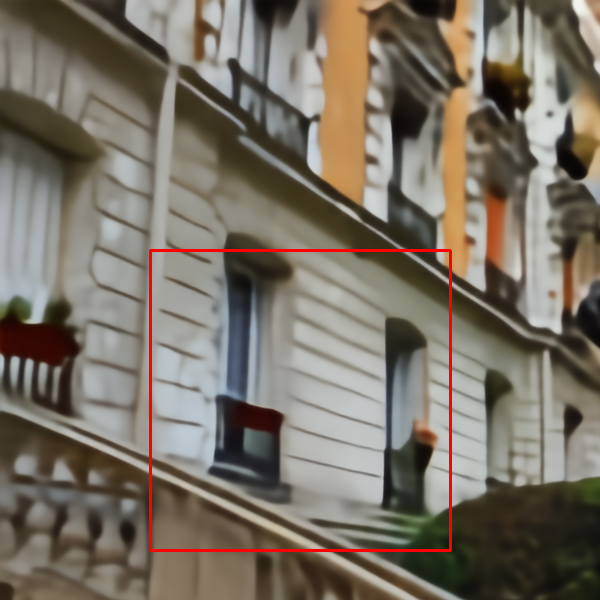}
        \caption*{Swin2SR~\cite{conde2022swin2sr}}
    \end{subfigure}%
    \begin{subfigure}{.33\linewidth}
        \centering
        \includegraphics[width=0.95\linewidth]{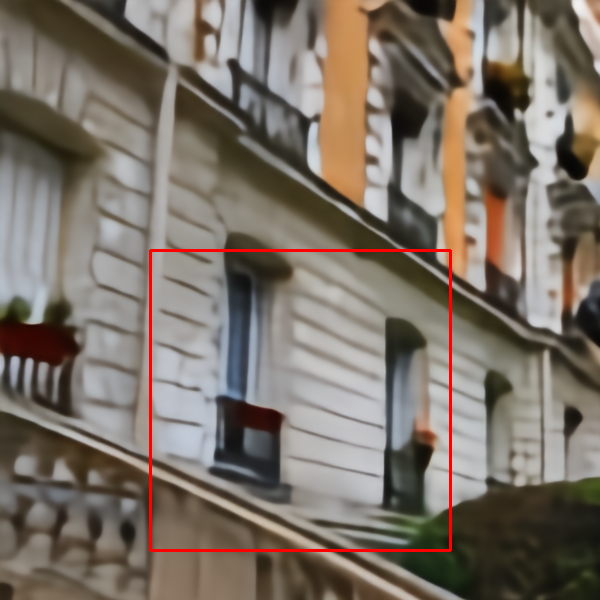}
        \caption*{MAXIM~\cite{tu2022maxim}}
    \end{subfigure}%
    \begin{subfigure}{.33\linewidth}
        \centering
        \includegraphics[width=0.95\linewidth]{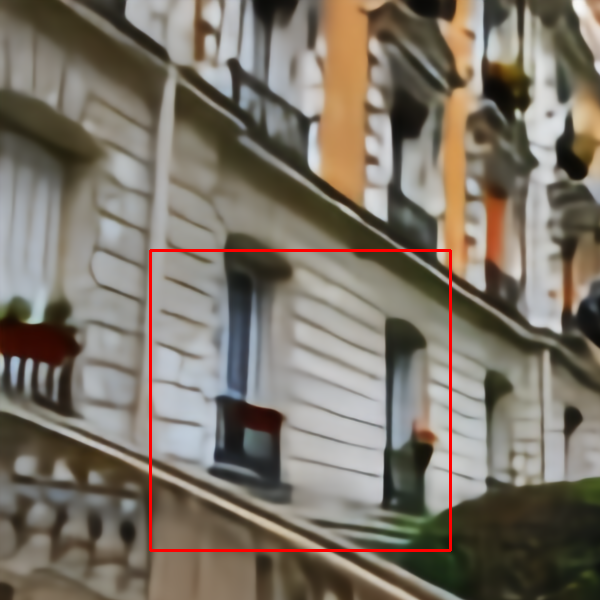}
        \caption*{AIRNet~\cite{Airnet}}
    \end{subfigure}
    \begin{subfigure}{.33\linewidth}
        \centering
        \includegraphics[width=0.95\linewidth]{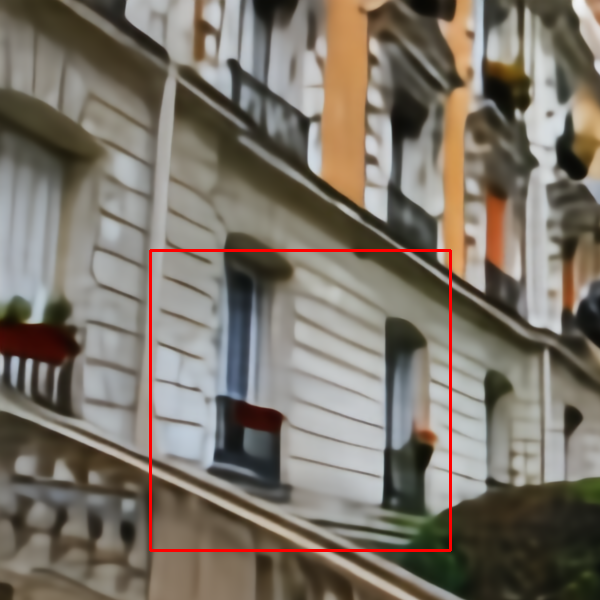}
        \caption*{PromptIR~\cite{promptir}}
    \end{subfigure}%
    \begin{subfigure}{.33\linewidth}
        \centering
        \includegraphics[width=0.95\linewidth]{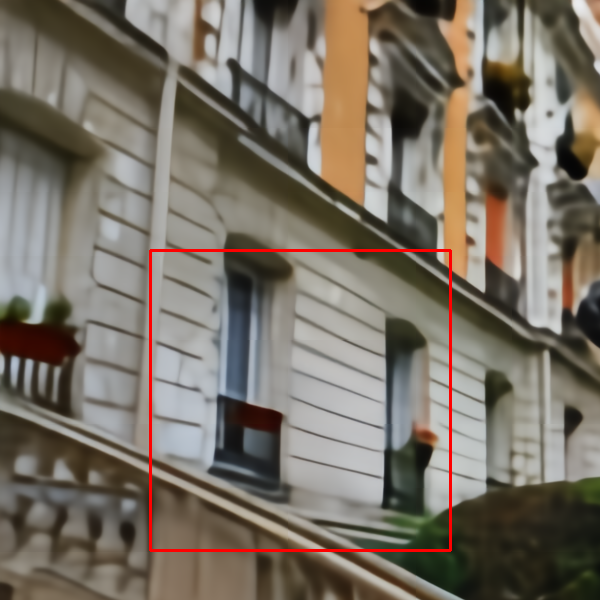}
        \caption*{UCIP}
    \end{subfigure}%
    \begin{subfigure}{.33\linewidth}
        \centering
        \includegraphics[width=0.95\linewidth]{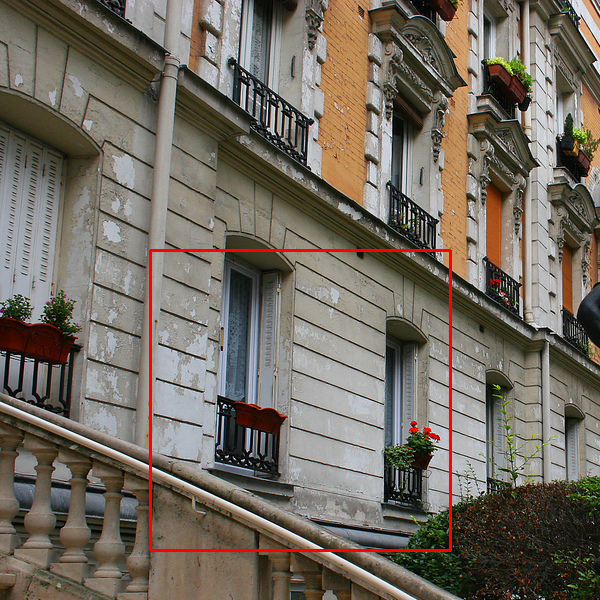}
        \caption*{HR}
    \end{subfigure}
    
    \caption{Visual Comparisons between UCIP and other methods on HIFIC ($\mathcal{Q}=\text{`high'}$) from Urban100~\cite{urban100}.}
\end{figure*}

\begin{figure*}
    \centering
    \begin{subfigure}{.33\linewidth}
        \centering
        \includegraphics[width=0.95\linewidth]{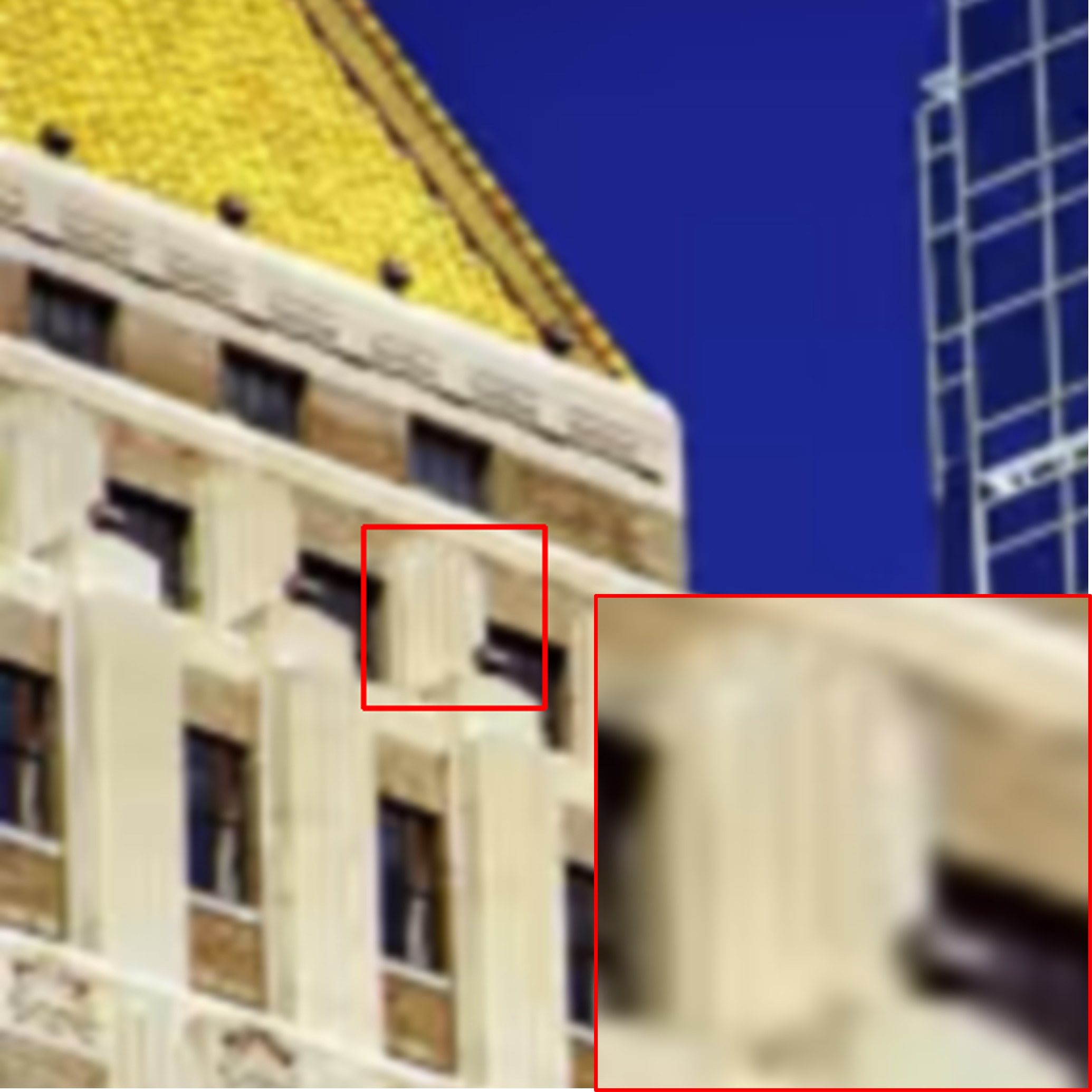}
        \caption*{Bicubic}
    \end{subfigure}%
    \begin{subfigure}{.33\linewidth}
        \centering
        \includegraphics[width=0.95\linewidth]{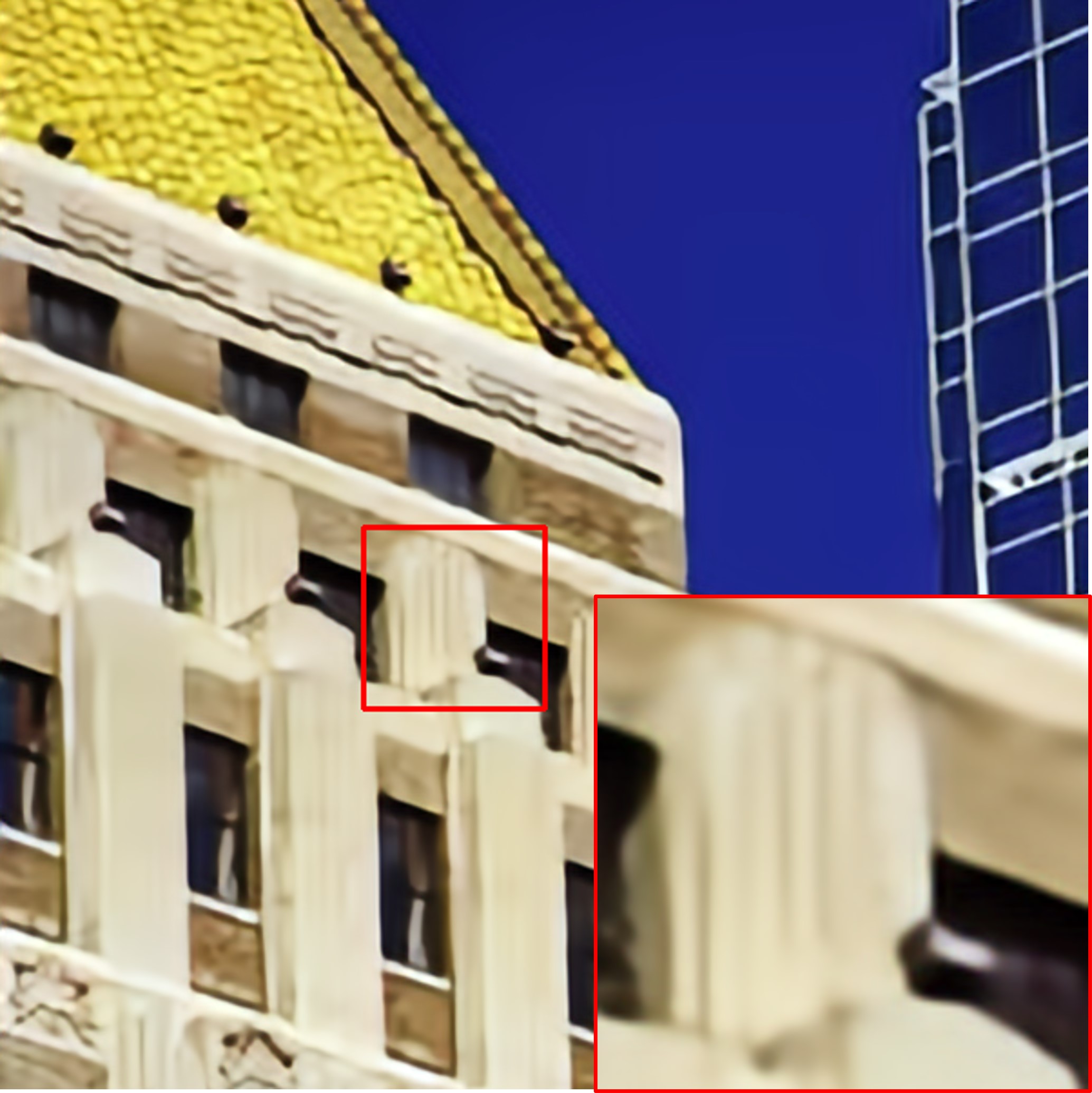}
        \caption*{RRDB~\cite{wang2018esrganRRDB}}
    \end{subfigure}%
    \begin{subfigure}{.33\linewidth}
        \centering
        \includegraphics[width=0.95\linewidth]{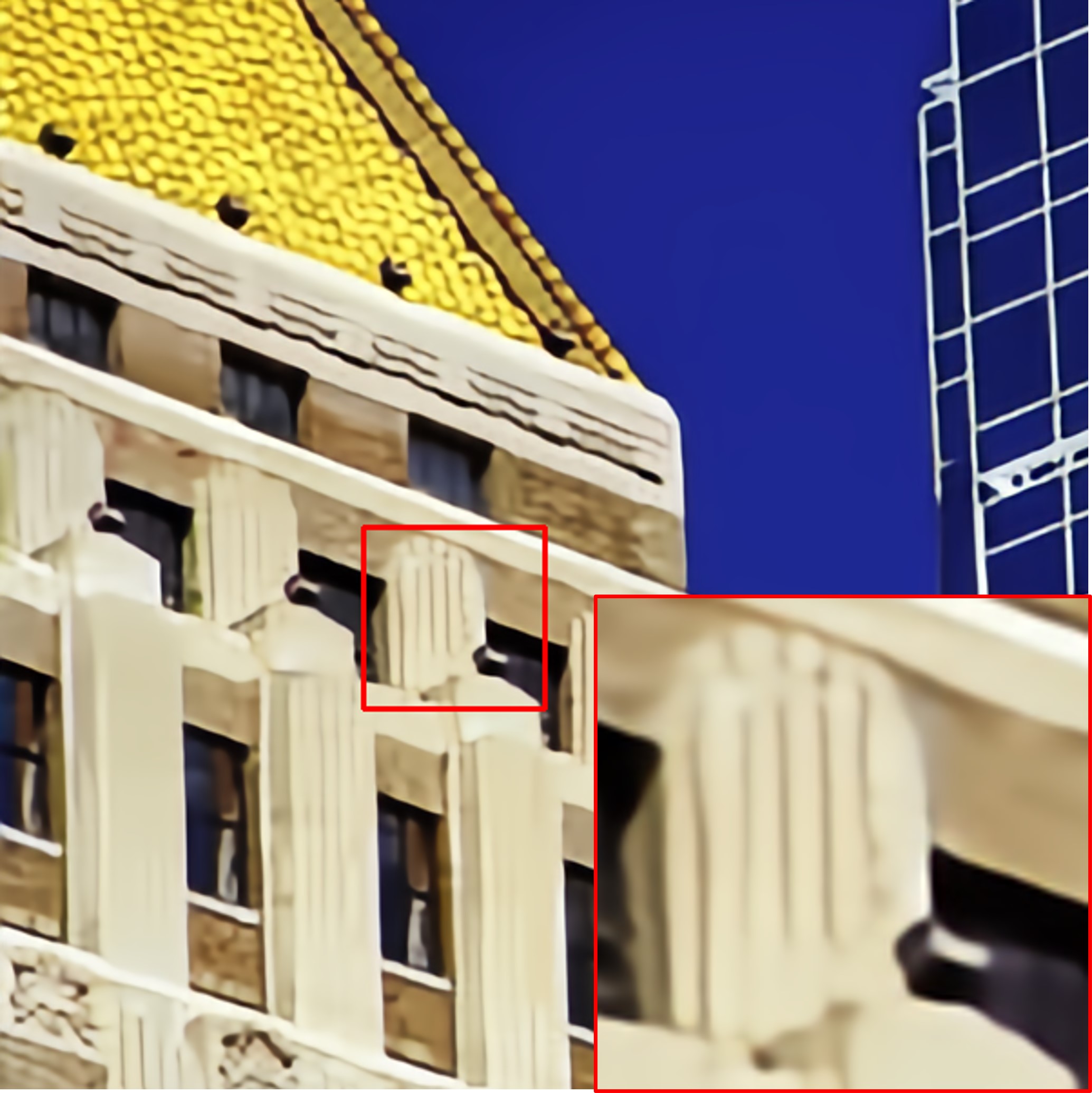}
        \caption*{SwinIR~\cite{liang2021swinir}}
    \end{subfigure}
    \begin{subfigure}{.33\linewidth}
        \centering
        \includegraphics[width=0.95\linewidth]{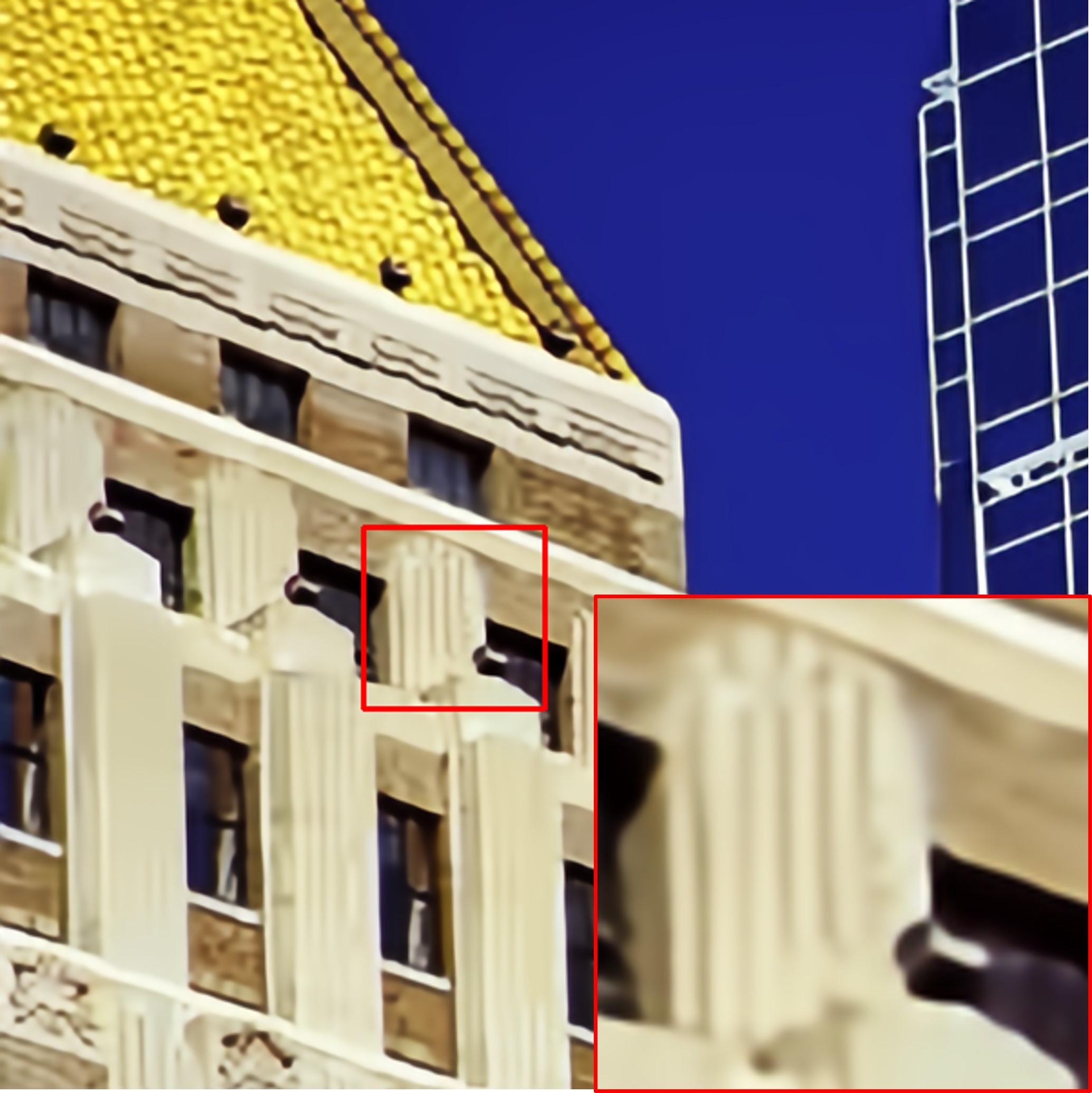}
        \caption*{Swin2SR~\cite{conde2022swin2sr}}
    \end{subfigure}%
    \begin{subfigure}{.33\linewidth}
        \centering
        \includegraphics[width=0.95\linewidth]{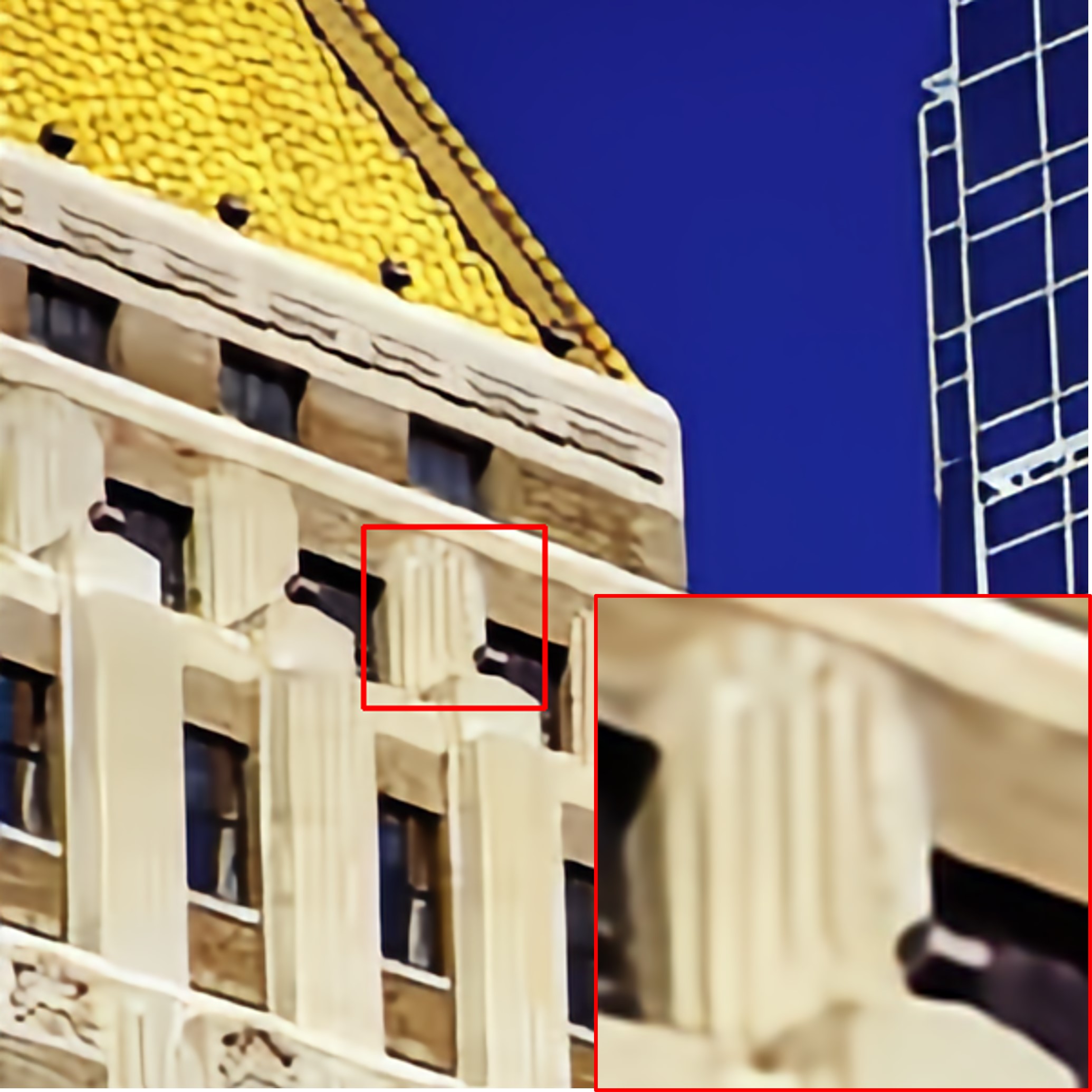}
        \caption*{MAXIM~\cite{tu2022maxim}}
    \end{subfigure}%
    \begin{subfigure}{.33\linewidth}
        \centering
        \includegraphics[width=0.95\linewidth]{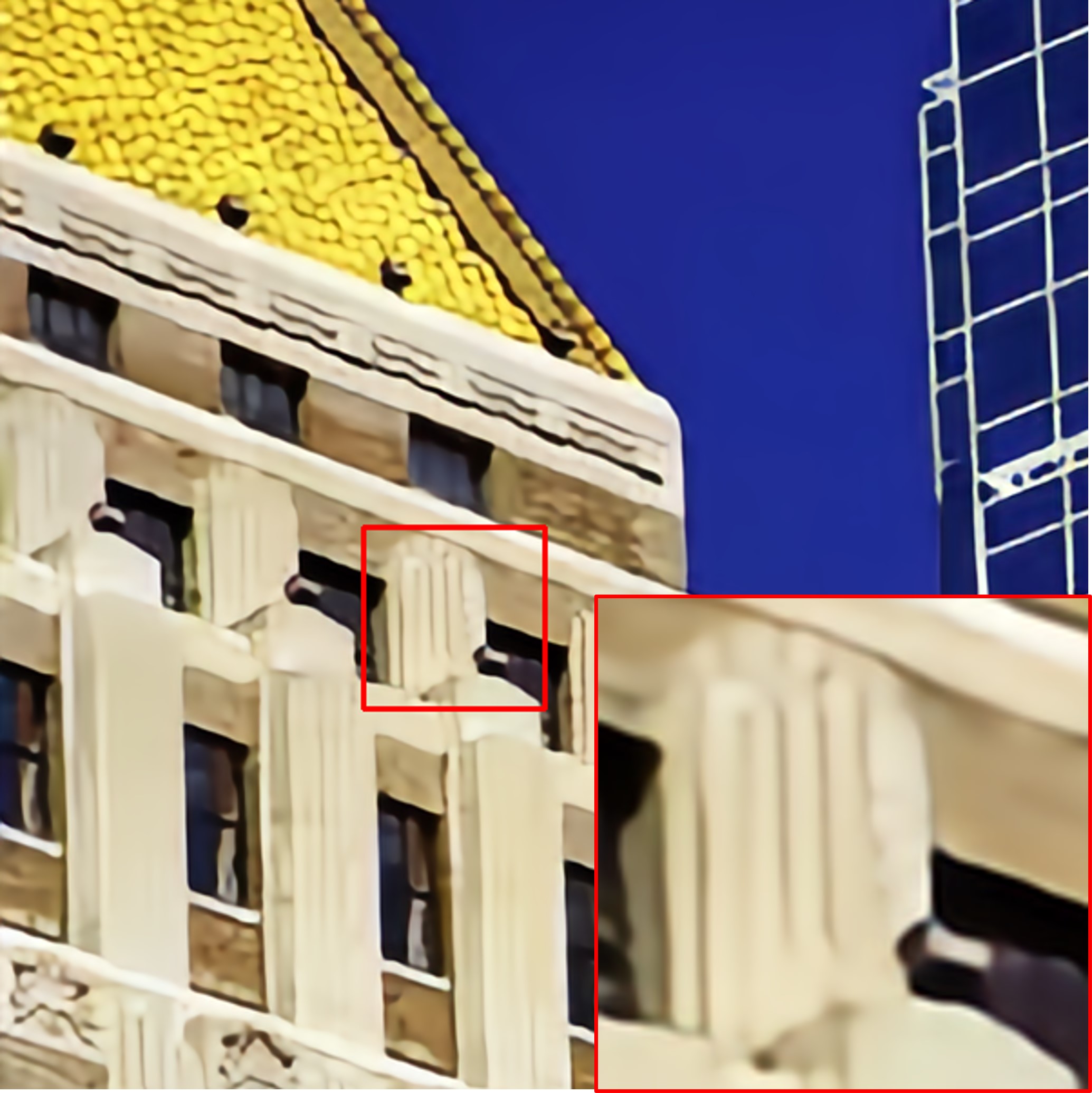}
        \caption*{AIRNet~\cite{Airnet}}
    \end{subfigure}
    \begin{subfigure}{.33\linewidth}
        \centering
        \includegraphics[width=0.95\linewidth]{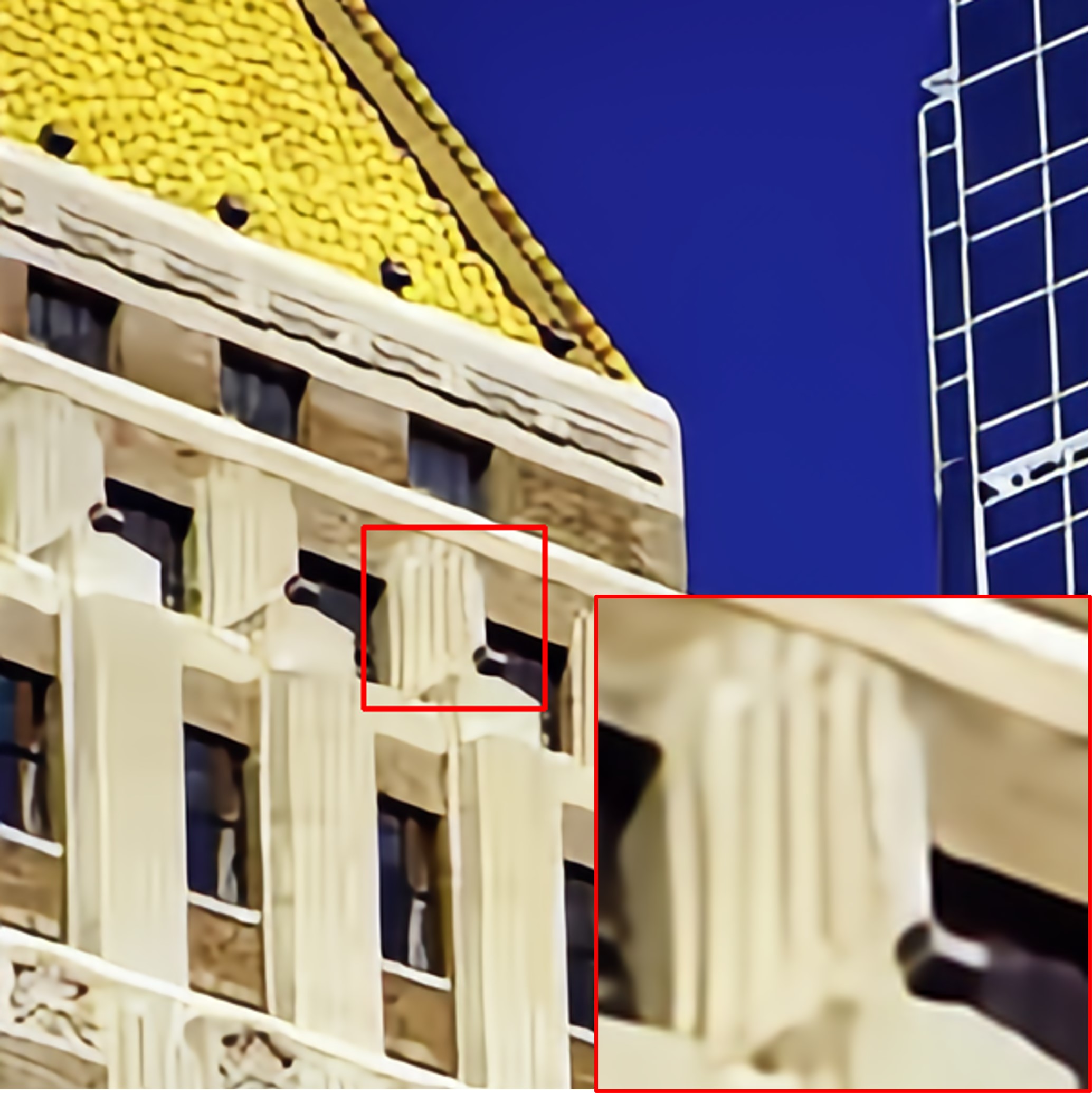}
        \caption*{PromptIR~\cite{promptir}}
    \end{subfigure}%
    \begin{subfigure}{.33\linewidth}
        \centering
        \includegraphics[width=0.95\linewidth]{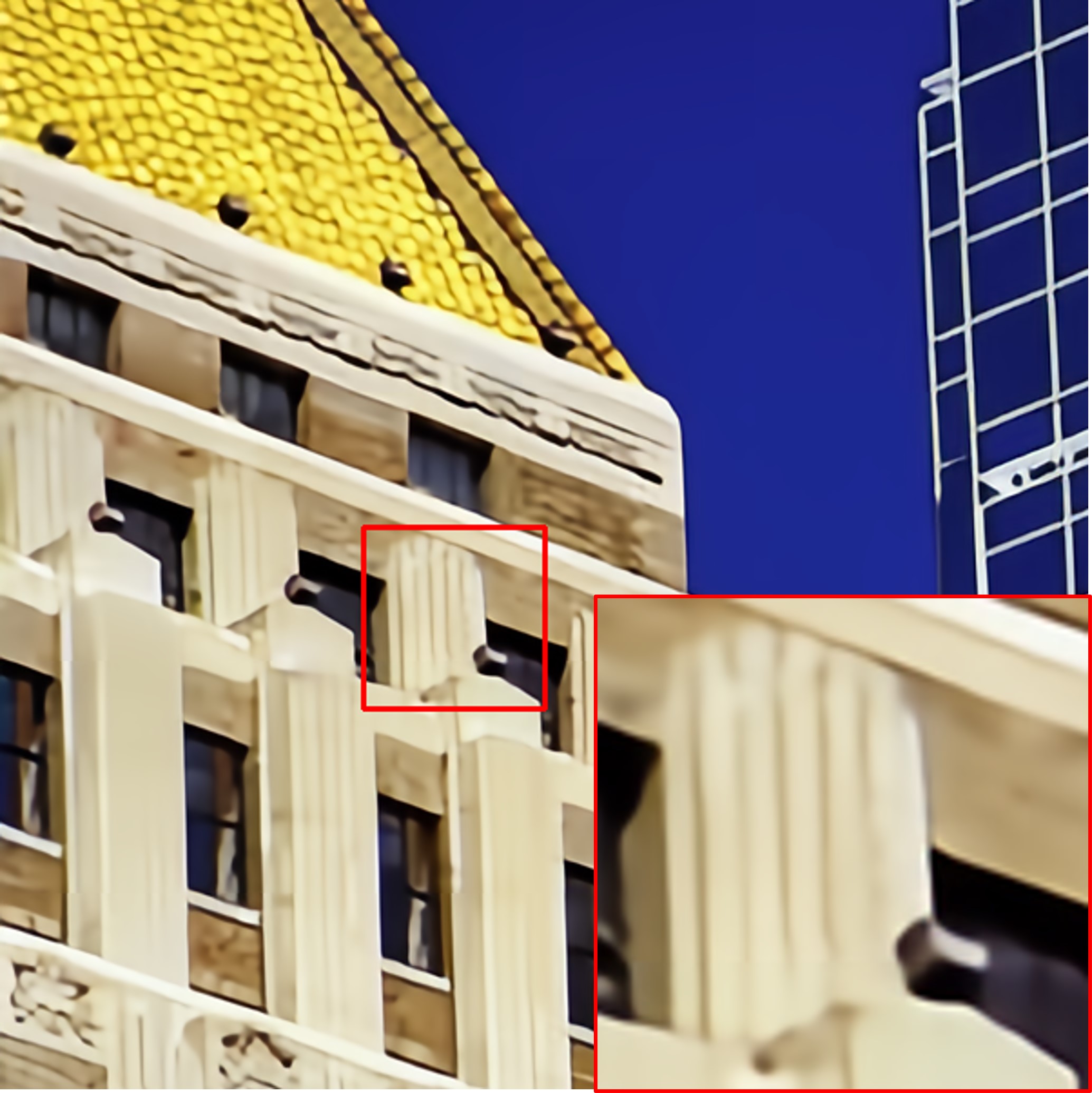}
        \caption*{UCIP}
    \end{subfigure}%
    \begin{subfigure}{.33\linewidth}
        \centering
        \includegraphics[width=0.95\linewidth]{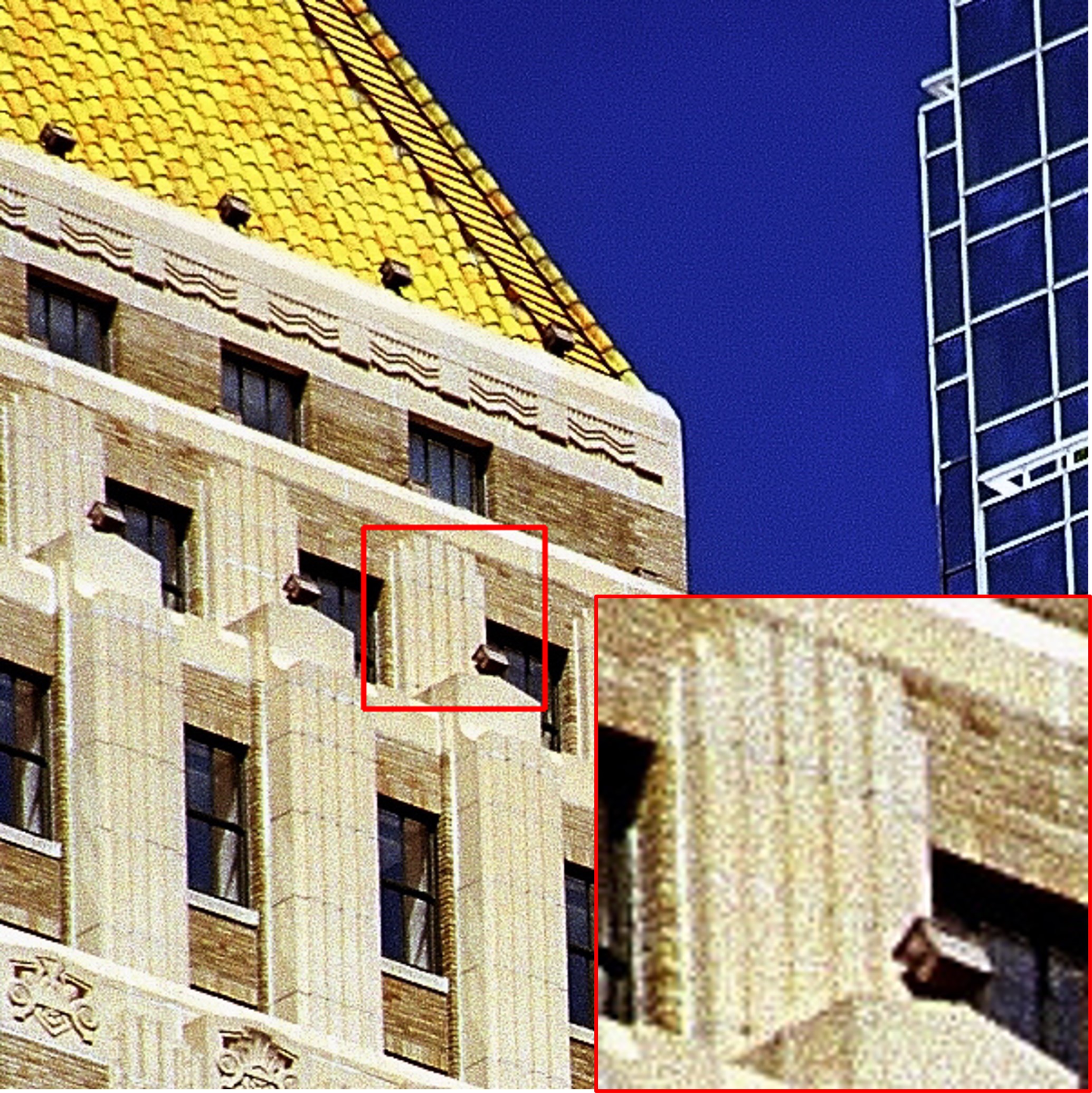}
        \caption*{HR}
    \end{subfigure}
    
    \caption{Visual Comparisons between UCIP and other methods on HM($\mathcal{Q}=32$) from Urban100~\cite{urban100}.}
\end{figure*}

\begin{figure*}
    \centering
    \begin{subfigure}{.33\linewidth}
        \centering
        \includegraphics[width=\linewidth]{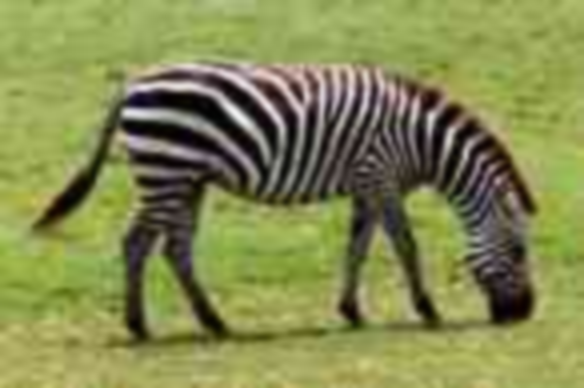}
        \caption*{Bicubic}
    \end{subfigure}%
    \begin{subfigure}{.33\linewidth}
        \centering
        \includegraphics[width=\linewidth]{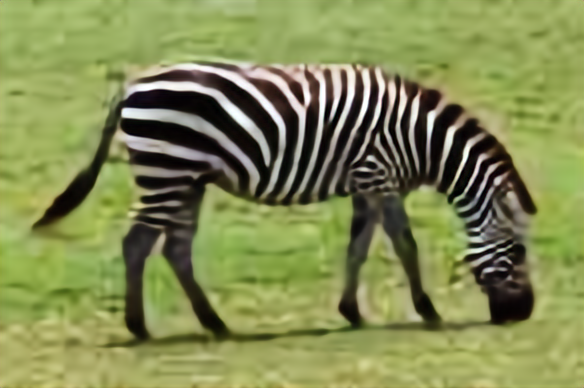}
        \caption*{RRDB~\cite{wang2018esrganRRDB}}
    \end{subfigure}%
    \begin{subfigure}{.33\linewidth}
        \centering
        \includegraphics[width=\linewidth]{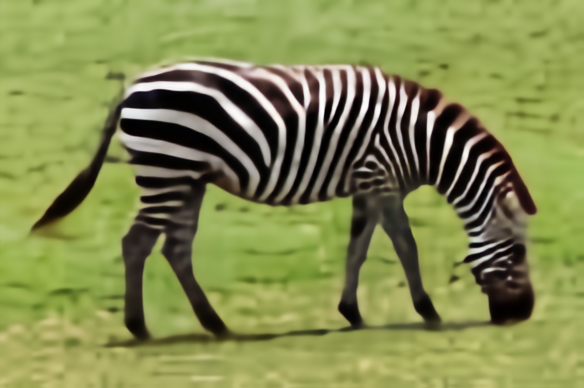}
        \caption*{SwinIR~\cite{liang2021swinir}}
    \end{subfigure}
    \begin{subfigure}{.33\linewidth}
        \centering
        \includegraphics[width=\linewidth]{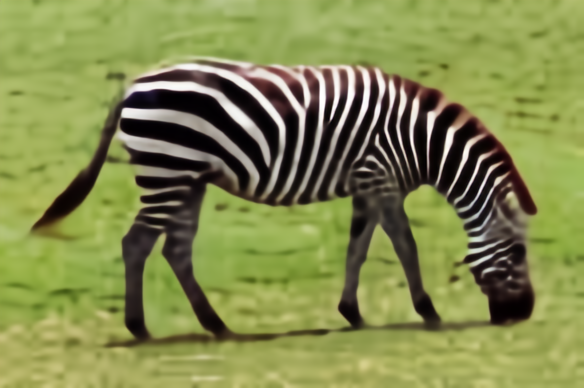}
        \caption*{Swin2SR~\cite{conde2022swin2sr}}
    \end{subfigure}%
    \begin{subfigure}{.33\linewidth}
        \centering
        \includegraphics[width=\linewidth]{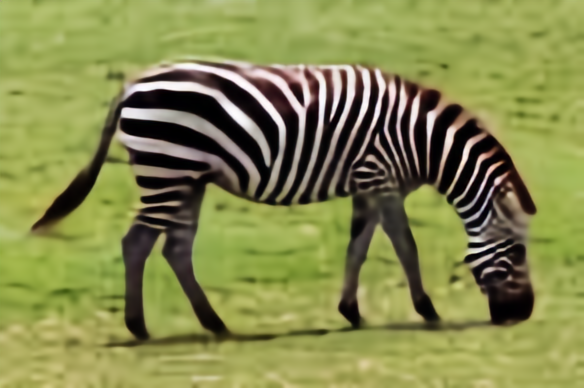}
        \caption*{MAXIM~\cite{tu2022maxim}}
    \end{subfigure}%
    \begin{subfigure}{.33\linewidth}
        \centering
        \includegraphics[width=\linewidth]{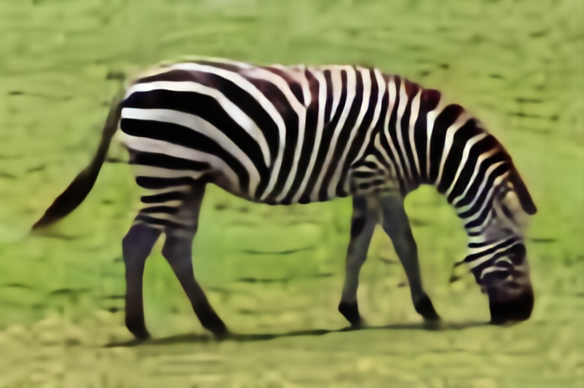}
        \caption*{AIRNet~\cite{Airnet}}
    \end{subfigure}
    \begin{subfigure}{.33\linewidth}
        \centering
        \includegraphics[width=\linewidth]{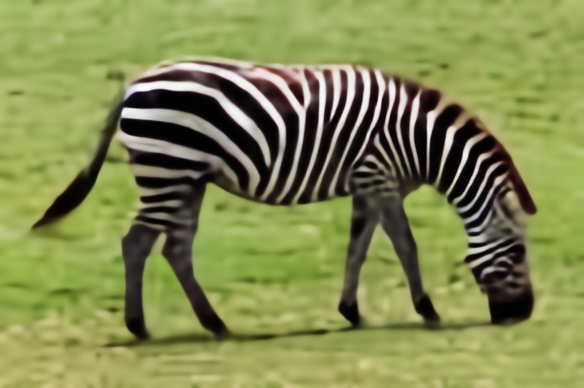}
        \caption*{PromptIR~\cite{promptir}}
    \end{subfigure}%
    \begin{subfigure}{.33\linewidth}
        \centering
        \includegraphics[width=\linewidth]{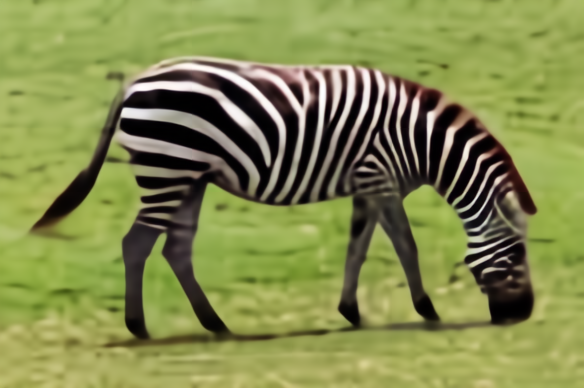}
        \caption*{UCIP}
    \end{subfigure}%
    \begin{subfigure}{.33\linewidth}
        \centering
        \includegraphics[width=\linewidth]{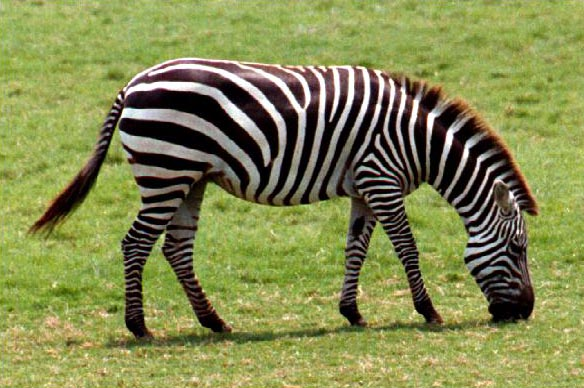}
        \caption*{HR}
    \end{subfigure}
    
    \caption{Visual Comparisons between UCIP and other methods on JPEG($\mathcal{Q}=30$) from Set14~\cite{Set14}.}
\end{figure*}

\begin{figure*}
    \centering
    \includegraphics[width=\linewidth]{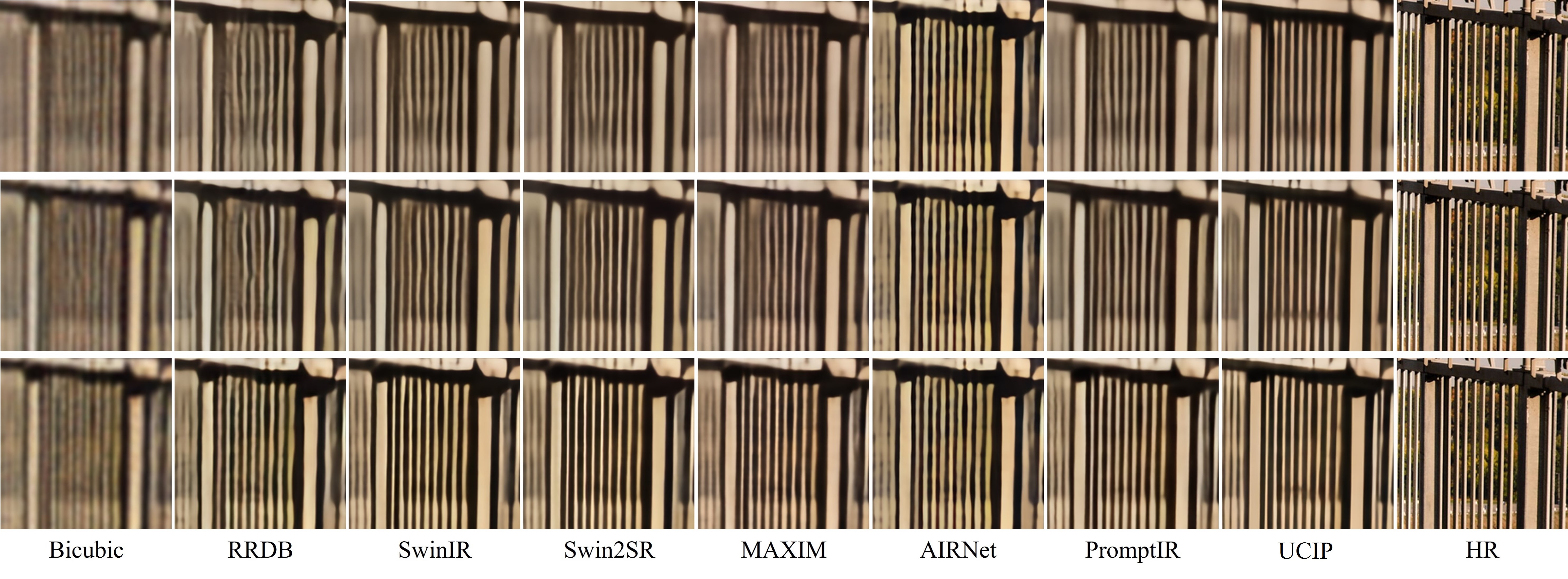}
    \caption{Visual comparisons between UCIP and other state-of-the-art methods under different compression qualities within HIFIC~\cite{mentzer2020hifi} codec. The qualities of HIFIC from top to bottom are `low', `medium', and `high', respectively.}
\end{figure*}

\end{document}